%% file: main.tex
\documentclass[twoside]{article}

\usepackage[accepted]{aistats2023}
\usepackage[round]{natbib}

\usepackage[utf8]{inputenc} % allow utf-8 input
\usepackage[T1]{fontenc}    % use 8-bit T1 fonts
\usepackage{hyperref}       % hyperlinks
\usepackage{url}            % simple URL typesetting
\usepackage{booktabs}       % professional-quality tables
\usepackage{amsfonts}       % blackboard math symbols
\usepackage{amsmath} 
\usepackage{nicefrac}       % compact symbols for 1/2, etc.
\usepackage{microtype}      % microtypography
\usepackage[dvipsnames]{xcolor}
\usepackage{tikz}
\usepackage{floatrow}
\usepackage{amssymb}
\usepackage{bbm}
\usepackage{dsfont}
\usepackage{mathtools}
\usepackage{enumitem}
\usepackage[font=small]{caption}
\usepackage[bottom]{footmisc}
\usepackage{algpseudocode}
\usepackage{algorithm}

\newtheorem{thm}{Theorem}
\newtheorem{Assumption}{Assumption}

\newcommand{\codecomment}[1]{\textbf{\color{black}// #1}}
\newcommand{\btheta}{\boldsymbol{\theta}}
\newcommand{\bTheta}{\boldsymbol{\Theta}}
\renewcommand{\hat}{\widehat}

\def\x{{\mathbf{x}}}
\newcommand\smallO{
  \mathchoice
    {{\scriptstyle\mathcal{O}}}% \displaystyle
    {{\scriptstyle\mathcal{O}}}% \textstyle
    {{\scriptscriptstyle\mathcal{O}}}% \scriptstyle
    {\scalebox{.7}{$\scriptscriptstyle\mathcal{O}$}}%\scriptscriptstyle
  }

\newcommand{\ann}[1]{\textbf{\color{purple}[Ann: #1]}}
\newcommand{\rafael}[1]{\textbf{\color{blue}[Rafael: #1]}}
\newcommand{\luca}[1]{\textbf{\color{olive}[Luca: #1]}}
\newcommand{\comment}[1]{}

\definecolor{awesome}{rgb}{1.0, 0.13, 0.32}
\definecolor{safetyorange}{rgb}{1.0, 0.4, 0.0}
\definecolor{vermilion}{rgb}{0.89, 0.26, 0.2}
\definecolor{aqua}{rgb}{0.0, 0.9, 0.9}
\newcommand{\add}[1]{{\color{black} #1}}

\newcommand{\remove}[1]{\textbf{\color{aqua}#1}}

\begin{document}

\runningtitle{Simulator-Based Inference with \textsc{Waldo}: Confidence Regions by Leveraging
Prediction Algorithms and Posterior Estimators}

\runningauthor{Luca Masserano, Tommaso Dorigo, Rafael Izbicki, Mikael Kuusela, Ann B. Lee}

\twocolumn[

\aistatstitle{Simulator-Based Inference with
\textsc{Waldo}: Confidence Regions by Leveraging
Prediction Algorithms and Posterior Estimators for Inverse Problems}

\aistatsauthor{
Luca Masserano\footnotemark[1]\textsuperscript{,}\footnotemark[2] 
\And 
Tommaso Dorigo\footnotemark[3] 
\And 
Rafael Izbicki\footnotemark[4] 
\And 
Mikael Kuusela\footnotemark[1]\textsuperscript{,}\footnotemark[2] 
\And 
Ann B. Lee\footnotemark[1]\textsuperscript{,}\footnotemark[2] 
}

\aistatsaddress{
\footnotemark[1]Department of Statistics \& Data Science, Carnegie Mellon University \\
\footnotemark[2]NSF AI Planning Institute for Data-Driven Discovery in Physics, Carnegie Mellon University \\
\footnotemark[3]INFN, Sezione di Padova, \quad 
\footnotemark[4]Department of Statistics, Federal University of S\~ao Carlos}

]

\begin{abstract}
Prediction algorithms, such as deep neural networks (DNNs), are used in many domain sciences to directly estimate internal parameters of interest in simulator-based models, especially in settings where the observations include images or 
%other 
complex high-dimensional data. In parallel, modern neural density estimators, such as normalizing flows, are becoming increasingly popular for uncertainty quantification, especially when both parameters and observations are high-dimensional. However, parameter inference is an inverse problem and not a prediction task; thus, an open challenge is to construct {\em conditionally valid} and {\em precise} confidence regions, with a guaranteed probability of covering the true parameters of the data-generating process, no matter what the (unknown) parameter values are, and without relying on large-sample theory. Many simulator-based inference (SBI) methods are indeed known to produce biased or overly confident parameter regions, yielding misleading uncertainty estimates. This paper presents \textsc{Waldo}, a novel method to construct
%for constructing 
confidence regions with finite-sample conditional validity by leveraging prediction algorithms or posterior estimators that are currently widely adopted in SBI. \textsc{Waldo} reframes the well-known Wald test statistic, and uses a computationally efficient regression-based machinery for classical Neyman inversion of hypothesis tests. We apply our method to a recent high-energy physics problem, where prediction with DNNs has previously led to estimates with prediction bias. We also illustrate how our approach can correct overly confident posterior regions computed with normalizing flows.
\end{abstract}

\section{INTRODUCTION}
\label{sec: intro}

%\add
{The vast majority of modern machine learning 
targets prediction problems, with 
algorithms such as Deep Neural Networks (DNNs) being particularly successful with point predictions
of a target variable $Y \in \mathbb{R}$
 when the input vectors $\mathbf{x} \in \mathcal{X}$
  represent complex high-dimensional data. In many science applications, however, one is often interested in the ``inverse'' problem of estimating the internal parameters of a data-generating process with reliable measures of uncertainty.  The parameters of interest, which we denote by   $\btheta$, are then not directly observed but are the “causes” of the observed data 
 $\mathbf{x}$.} 
{In order to make inference on internal parameters, one needs a statistical model that relates the (unknown) parameters to the observed data. In science and engineering, 
simulations are often used to model the behavior of complex systems in lieu of an analytical likelihood, when the latter is too complicated to be evaluated explicitly.
%deemed to  oversimplify the data-generating process or not faithfully describe actual observations. 
Let $\mathcal{D} \coloneqq (\mathbf{x}_1, \dots, \mathbf{x}_n)^T$ denote observable data, where the ``sample size'' $n$ refers to the number of observations at a fixed configuration of the parameters $\btheta$. {\em Likelihood-free inference} (LFI), which is a form of simulator-based inference (SBI; \cite{cranmer2020frontier}), refers to parameter estimation in a setting where the likelihood function $\mathcal{L}(\btheta;\mathcal{D}) \coloneqq p(\mathcal{D|\btheta})$ itself is intractable, but the scientist, in lieu of an explicit likelihood, has access to a %\remove{stochastic} 
simulator that can generate $\mathcal{D}$ given any $\mathbf{\btheta} \in \bTheta$.}

\iffalse 
% commenting previous text
\remove{Simulations can be used to predict how systems will behave in a variety of circumstances, and can help to constrain parameters when Gaussian 
or other parametric likelihood models become questionable at the level of precision of modern scientific data.
Stochastic simulators are able to produce observable data $\mathbf{x}$ under different parameter settings, but they encode the likelihood function $\mathcal{L}(\btheta;\mathbf{x}) \coloneqq p(\mathbf{x|\btheta})$ only implicitly. Statistical inference when $\mathcal{L}(\btheta;\mathbf{x})$ is intractable is often dubbed likelihood-free inference (LFI), and is a form of simulation-based inference (SBI).}
\fi

LFI has undergone a revolution in terms of the complexity of problems that can be tackled, both because of faster and more realistic simulators that can generate a large number of examples $\mathcal{T}=\{ (\btheta^{(j)}, \mathcal{D}^{(j)}) \}_{j=1}^{B}$, and because of more powerful AI techniques that can learn various quantities of interest from these  simulations.  DNNs -- such as convolutional neural networks (CNNs) \citep{lecun1995convolutional} -- are now used in many domain sciences to directly {\em predict} internal parameters of interest in statistical models, especially in settings where $\x$ represents images or other high-dimensional data. Recent examples include estimating the energy ($\theta$) of muons that radiate photons when traversing a finely segmented calorimeter ($\x$) \citep{kieseler2022calorimetric};  estimating the mass of a galaxy cluster ($\theta$) from velocities and projected radial distances  ($\x$) for a particular line-of-sight of the observer relative to the galaxy cluster \citep{ho2019robust}; and estimating the range and noise-to-signal covariance parameters ($\btheta$) of spatial Gaussian processes from spatial fields or variograms ($\mathbf{x}$) \citep{gerber2021fast}. In parallel, modern neural density estimators, such as normalizing flows, are becoming increasingly popular for uncertainty quantification, especially when both parameters $\btheta$ and observations $\x$ are high-dimensional. Recent examples include \citet{boyda2021sampling,mishra2022neural,lueckmann2021benchmarking}.
 
%\add
%DNNs are often easier to train than neural density estimators, 
%\tommaso {is this a safe claim or do we need to justify it? DNNs can be real bitches} but 
Purely predictive approaches are known to suffer from prediction bias in inverse problems, as the point prediction -- e.g., $\mathbb{E}[\btheta|\x]$ under squared error loss -- is generally different from the true (unknown) parameter $\btheta$. Concrete examples include \citet{dorigo2022deep, ho2019robust, kiel2019bias}, where attempts are made to correct for the observed bias post-hoc. 
%The problem of producing reliable uncertainty estimates of internal parameters from DNN point predictions has also not yet found a solution. 
At the same time, many posterior estimation methods are known to be overly confident, meaning that they yield confidence sets with empirical coverage lower than the desired nominal level \citep{hermans2021averting}, hence leading to potentially misleading results.

At the heart of the matter is the fact that both  predictive  and posterior approaches in SBI rely heavily on how the values of $\btheta$ in the training set $\mathcal{T}$ are sampled. For reliable inference, however, the coverage guarantees of the confidence sets should be independent of the choice of prior $\pi_{\btheta}$, thereby allowing the user to design priors that can lead to tighter, {\em but guaranteed to be valid}, confidence sets.
In this work, we present a solution without relying on large-sample theory or computationally intensive Monte Carlo sampling.
 
\begin{figure*}[t!]
    %\floatbox[{\capbeside\thisfloatsetup{capbesideposition={right,center}}}]{figure}[\FBwidth]
    {\caption{
    \textbf{Schematic diagram of \textsc{Waldo}.} \add{{\em Left (\textcolor{RoyalBlue}{blue})}: For a first simulated set $\mathcal{T}$, we estimate the conditional mean $\mathbb{E}[\btheta|\mathcal{D}]$ and variance $\mathbb{V}[\btheta|\mathcal{D}]$ using a prediction algorithm (e.g., DNN) or posterior estimator (e.g., normalizing flows). This gives us the \textsc{Waldo} test statistic $\hat{\tau}^{\textsc{Waldo}}$ in Equation~\ref{eq: waldo_statistic}.}
    {\em Center (\textcolor{PineGreen}{green}):}  \add{For a second simulated set $\mathcal{T}^{\prime}$}, we estimate critical values $\hat{C}_{\btheta_0,\alpha}$ for all tests $H_0: \btheta = \btheta_0$ across the parameter space $\bTheta$ via a quantile regression of $\hat{\tau}^{\textsc{Waldo}}$ on $\btheta$. {\em Bottom:} \add{Given an observation $D$, Neyman inversion converts the tests (which compare test statistics with critical values) into a confidence region for $\theta$.} {\em Right (\textcolor{RedOrange}{red}):} \add{For a third simulated set~$\mathcal{T}^{\prime\prime}$, we provide an independent assessment of the conditional validity of constructed confidence regions by computing coverage diagnostics across the entire parameter space.} See Section~\ref{sec: waldo_methodology} \add{and Algorithm~\ref{algo:waldo_algo}} for details.}
    
    % ALTERNATIVE CAPTION FROM WORKSHOP PAPER
    %\caption{\textbf{Schematic diagram of \textsc{Waldo}.} The Neyman construction of confidence sets (\textit{bottom}) requires estimation of a test statistic $\tau^{\textsc{Waldo}}$ (\textit{left}), and of critical values $C_{\btheta_0, \alpha}$ across the whole parameters space (\textit{center}). Diagnostics (\textit{right}) are then used to make sure that those confidence regions achieve the desired level of conditional coverage. See Section~\ref{sec: waldo_methodology} and Algorithm~\ref{algo: waldo_algo} for details. }
    \label{fig:waldo_framework}}
    {\includegraphics[width=0.8\textwidth]{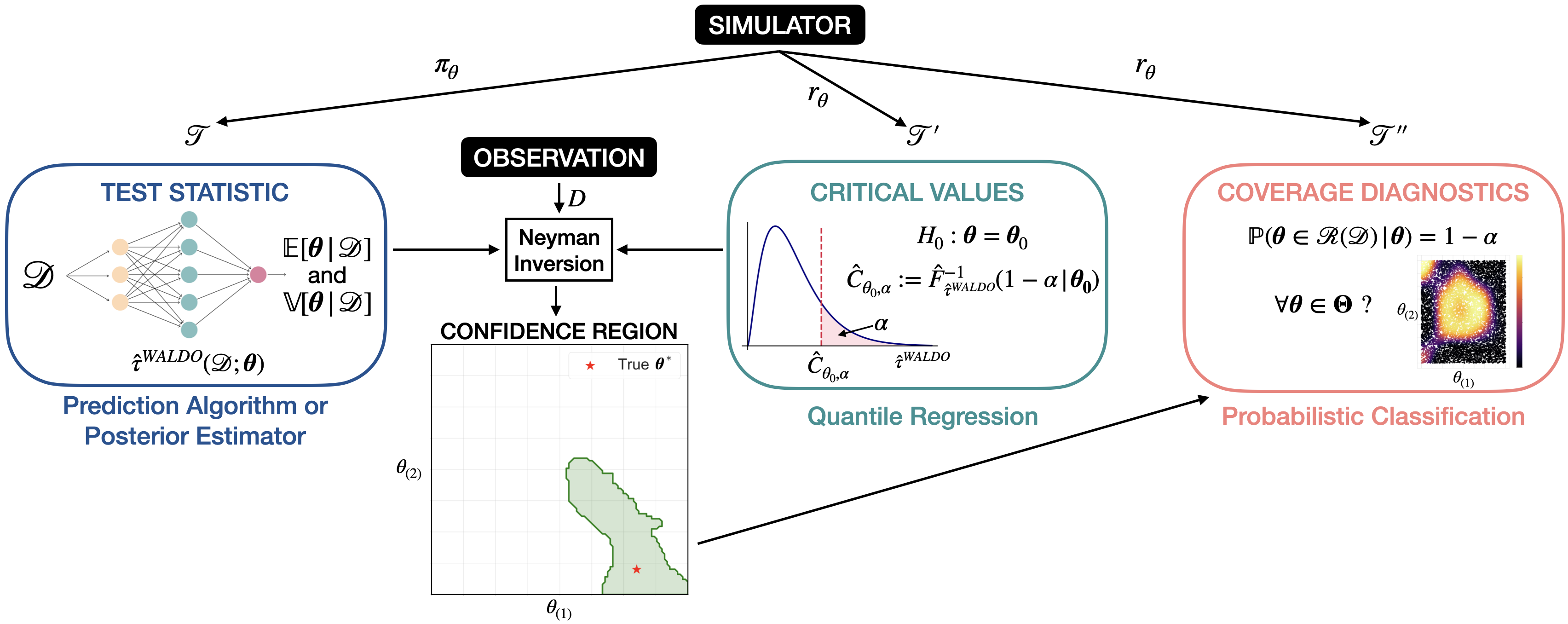}}
\end{figure*}

\textsc{Waldo} is a new LFI procedure that can leverage any %\remove{high-capacity}
prediction algorithm or neural posterior estimator to construct confidence regions for $\btheta$ with correct \emph{conditional coverage}; that is,
%\remove{parameters regions} 
sets $\mathcal{R}(\mathcal{D})$ satisfying
\begin{equation}
\label{eq:conditional_coverage}
\mathbb{P}(\btheta \in \mathcal{R}(\mathcal{D})|\btheta) = 1-\alpha , \quad \forall \btheta \in \bTheta,
\end{equation}
regardless of the size $n$ of the observed sample, where $(1-\alpha) \in (0,1)$ is a prespecified confidence level. 
% confidence % miscoverage %\tommaso {as an ignorant, this way of calling alpha is a bit confusing to me - alpha is a constant set before the inference machinery is even set up - to me, calling it miscoverage hides its role} 
Correct conditional coverage implies correct \emph{marginal coverage}, $\mathbb{P}(\btheta \in \mathcal{R}(\mathcal{D})) = 1 - \alpha$, but the former is a stronger requirement that checks that the confidence set is calibrated no matter what the true parameter is, whereas marginal coverage only requires the set to be calibrated on average over the parameter space~$\bTheta$.
%\luca{I moved the below description here. It makes more sense to have it before diagnostics.}
\textsc{Waldo} reframes the Wald test \citep{wald1943tests} and leverages existing prediction or posterior algorithms to first compute a test statistic (Equation~\ref{eq: waldo_statistic}) based on estimates of the conditional mean $\mathbb{E}[\btheta|\mathcal{D}]$ and conditional variance $\mathbb{V}[\btheta|\mathcal{D}]$. It then uses a recent approach \citep{dalmasso2021likelihood} to the Neyman construction \citep{neyman1937outline}, which estimates critical values via quantile regression and converts hypothesis tests into a confidence region with finite-$n$ conditional coverage. \textsc{Waldo} also includes an independent diagnostics module to check that the constructed confidence sets achieve the correct nominal level of empirical coverage across the parameter space. Section \ref{sec: waldo_methodology} describes our methodology in detail, and Figure~\ref{fig:waldo_framework} summarizes its different components.

\textsc{Waldo} embraces the best sides of both the Bayesian and frequentist perspectives to statistical inference by providing confidence sets that \textit{(i)} can effectively exploit available domain-specific knowledge, further constraining parameters when the prior is consistent with the data, and \textit{(ii)} are guaranteed to have the nominal conditional coverage even in finite samples as long as the quantile regressor is well estimated, regardless of the correctness of the prior. \textsc{Waldo} is also amortized, meaning that once the procedure has been trained, it can be evaluated on any number of observations. \add{We lay out the statistical and computational properties of \textsc{Waldo}, providing synthetic examples with analytical solutions to verify and support our claims (see Section~\ref{sec: waldo_stat_properties} and Section~\ref{sec: comput_properties}). We then show its effectiveness on two complex applications, which confirm the results we obtained on the synthetic examples: the first one (Section~\ref{sec: mixture_bayesianLFI}) uses an established benchmark in SBI and leverages posterior distributions to construct valid confidence sets regardless of the prior distribution. The second application (Section~\ref{sec: muons}) deals with a current problem in high-energy physics: inferring the energy of muons from a particle detector exploiting predictions from a custom CNN and an innovative source of information, i.e., the pattern of energy deposits left by muons in a finely segmented calorimeter. The results we obtain for this problem are of scientific interest by themselves, as a rigorous estimate of the uncertainty around estimated muon energies is essential in the search of new physics. A ready-to-use and flexible implementation of \textsc{Waldo} is available at \href{https://github.com/lee-group-cmu/lf2i}{\texttt{https://github.com/lee-group-cmu/lf2i}}.}

\paragraph{Notation} We refer to parameters of interest as $\boldsymbol{\theta} \in \boldsymbol{\Theta} \subset \mathbb{R}^p$ and to a sample of size $n$ of observable input data as $\mathcal{D} = (\mathbf{x}_1, \dots, \mathbf{x}_n)^T$, with $\mathbf{x}_i \in \mathcal{X} \subset \mathbb{R}^{d}$ and possibly $p \neq d$. Note that $n$ is distinct from $B, B^\prime$ and $B^{\prime\prime}$, i.e., the number of simulations required at different steps of our method. We distinguish between observable data and actual observations by denoting the latter as $D$. We refer to confidence regions as $\mathcal{R}(\mathcal{D})$. The terms ``set'', ``region'' and (when $p=1$) ``interval'' are used interchangeably.

\section{RELATION TO OTHER WORK}
\label{sec: related_work}
\add{There exist many approaches for calibrating predictive distributions $p(y| \x)$ to achieve marginal or conditional validity in ``forward'' $\x \rightarrow y$ problems; examples include conformal inference \citep{vovk2005algorithmic,lei2018distribution,chernozhukov2021distributional} and the calibration procedures of \citet{bordoloi2010photo,dey2022calibrated}.
In the Bayesian inference domain, such calibration procedures correspond to ensuring that an estimate $\hat{p}(\theta | \x)$ of the posterior $p(\theta |\x)$  indeed corresponds to the true posterior  implied by the prior that was used.  This work, on the other hand, deals with the question of constructing {\em confidence sets} with correct conditional coverage for internal unknown parameters $\theta$ in so-called ``inverse problems'' (recall Equation \ref{eq:conditional_coverage}), which is not the same as achieving conditional coverage for prediction sets, or recalibrating posteriors.}

Similarly, existing approaches for deep learning uncertainty quantification (see \cite{gawlikowski2021survey} for a recent review), such as Monte Carlo drop out \citep{Gal2016} and conformal inference DNNs \citep{papadopoulos2007conformal}, construct prediction sets instead of confidence sets. Before \textsc{Waldo}, there has been no straightforward way to obtain confidence sets from point predictions or estimated posteriors obtained from deep neural networks and other predictive ML algorithms.

\add{For example,} various domain science applications have developed post-hoc corrections to predictive or posterior inferences to reduce observed biases and to improve the calibration of uncertainties. \add{Such corrections are common in areas ranging} from particle physics \citep{dorigo2022deep} to cosmology \citep{ho2019robust} and remote sensing \citep{kiel2019bias}. Usually the goal of the corrections is to reduce the impact of the prior specification, but in contrast to \textsc{Waldo}, \add{post-hoc correction} approaches do not provide formal coverage guarantees. \add{Similarly, in some settings, priors can be designed so that credible regions achieve correct conditional coverage  \citep{Bayarri2004,Berger2006,Kass1996,Scricciolo1999,Datta2005}. However, this technique requires knowledge of the likelihood function (which is not available in LFI). Moreover, such  prior distributions often do not encode  actual prior information, a limitation that is not present in  \textsc{Waldo}.}

\add{Finally,} posterior inferences do not control conditional coverage even for correctly specified priors \citep{patil2020objective}. \textsc{Waldo} addresses \add{this problem using Neyman inversion via an efficient regression-based approach proposed in \cite{dalmasso2021likelihood}. In the latter work, however, the authors construct likelihood-based test statistics (the Bayes factor or likelihood ratio)  which require an extra numerical integration or optimization step that can lead to a loss of power of the resulting confidence sets. \textsc{Waldo}, on the other hand, has the ability of directly leveraging flexible prediction algorithms and posterior estimators to construct valid and potentially more precise finite-$n$ confidence sets.}

\section{METHODOLOGY}
\label{sec: methodology}

\add{\textsc{Waldo} leverages a regression-based approach to the Neyman construction, reframing the Wald test to use the output of common LFI prediction algorithms and posterior estimators. After outlining its statistical foundations, we describe our procedure and its properties using synthetic examples.}

\subsection{Foundational Tools from Classical Statistics}
\label{sec: waldo_foundations}

\paragraph{Neyman construction} 
A key ingredient of \textsc{Waldo} is the equivalence between hypothesis tests and confidence sets, which was formalized by \cite{neyman1937outline}. The \add{basic} idea is to invert a series of level-$\alpha$ hypothesis tests of the form 
\begin{equation}
\label{eq: hypothesis_test}
H_0: \btheta = \btheta_0 \quad \mathrm{vs.} \quad H_1: \btheta \neq \btheta_0,
\end{equation} for all $\btheta_0 \in \bTheta$. After observing a sample $D$, %\remove{we then define} 
\add{one constructs} 
a confidence region $\mathcal{R}(D)$ for $\btheta$ by taking all $\btheta_0$ \add{values} that were not rejected by the \add{series of} tests above. By \add{design}, the set $\mathcal{R}(\mathcal{D})$ satisfies Equation~\eqref{eq:conditional_coverage}, i.e., \add{it} has the correct \add{$1-\alpha$} coverage \add{across the \textit{entire} parameter space $\bTheta$}. Albeit simple, the Neyman construction \add{requires one to control the type I error for every $\theta \in \Theta$}. It is therefore hard to implement in practice \add{within an LFI setting, without resorting to large-$n$ approximations like Wilks' theorem \citep{wilks1938LRAsymptotic}, or to Monte Carlo approaches, which become computationally prohibitive as the dimensionality of the parameter space increases (\cite{cousins2018lectures}; see also Section~\ref{sec: comput_properties}). }
\comment{ % OLD TEXT
to estimate the critical values $C_{\btheta_0, \alpha}$ at fixed $\theta_0$-values in $\Theta$. \remove{since it requires estimating critical values $C_{\btheta_0, \alpha}$ that define the level-$\alpha$ acceptance region for {\em every} hypothesis test that we invert. In the context of LFI, one usually either resorts to asymptotic results, which are not valid when the data generating process is irregular or the sample size $n$ is small \citep{algeri2019searching, lyons2018statistical, cowan2011asymptotic}, or Monte Carlo approaches \citep{mackinnon2009bootstrap, ventura2010bootstrap}, which become computationally prohibitive as the dimensionality of the parameter space increases \citep{cousins2018lectures}.}
}

\paragraph{Wald test} \add{Since any test that controls the type I error at level $\alpha$ can be used for the Neyman construction, we base \textsc{Waldo} on the classical Wald test \citep{wald1943tests}, which is uniformly most powerful in many settings \citep{ghosh1991higher,lehmann2005testing}. The Wald test measures the agreement of the data with the null hypothesis for $\btheta$, and it has the following form for $p=1$:}
\begin{equation}
\label{eq: wald_statistic}
\tau^{\textsc{Wald}}(\mathcal{D};\theta_0) \coloneqq \frac{(\hat{\theta}^{\textsc{MLE}} - \theta_0)^2}{\mathbb{V}(\hat{\theta}^{\textsc{MLE}})},
\end{equation}
where $\hat{\theta}^{\textsc{MLE}}$ is the maximum-likelihood estimator of $\theta$ and $\hat{\mathbb{V}}(\hat{\theta}^{\textsc{MLE}})$ can be any consistent estimator of its variance. However, in our setting, we do not have access to the likelihood and we cannot resort to assumptions on the distribution of $\tau^{\textsc{Wald}}(\mathcal{D};\theta_0)$, nor to asymptotic regimes, which makes it difficult to directly compute the Wald test statistic.

\subsection{Confidence Sets from Predictions and Posteriors}
\label{sec: waldo_methodology}

\paragraph{From Wald to \textsc{Waldo}} \textsc{Waldo} \add{reframes the Wald test by} replacing $\hat{\theta}^{\textsc{MLE}}$ and its variance with \add{quantities that are easily computable with prediction algorithms or posterior estimators commonly used in LFI.} \add{We define the \textsc{Waldo} test statistic for parameters of arbitrary dimensionality $p$ as} 
\begin{equation}
\label{eq: waldo_statistic}
    \tau^{\textsc{Waldo}}(\mathcal{D};\btheta_0) = (\mathbb{E}[\btheta|\mathcal{D}] - \btheta_0)^T \mathbb{V}[\btheta|\mathcal{D}]^{-1} (\mathbb{E}[\btheta|\mathcal{D}] - \btheta_0),
\end{equation}
where $\mathbb{E}[\btheta|\mathcal{D}]$ and $\mathbb{V}[\btheta|\mathcal{D}]$ are, respectively, the conditional mean and covariance matrix of $\btheta$ given $\mathcal{D}$. \add{The connection to the Wald test follows from the asymptotic behavior of Bayes estimators (e.g., \cite{chao1970asymptotic, ghosh2003preliminaries, ghosh1982expansions, li2020deviance}):}
\begin{align*}
    &\mathbb{E}[\btheta|\mathcal{D}] - \hat{\btheta}^{\textsc{MLE}} = \smallO_p(n^{-1/2}) \quad \text{and} \quad \\
    &\mathbb{V}[\btheta|\mathcal{D}] - \frac{1}{n}H^{-1}(\hat{\btheta}^{\textsc{MLE}}) = \smallO_p(n^{-1}),
\end{align*}
where $H^{-1}(\hat{\btheta}^{\textsc{MLE}})$ is the negative inverse Fisher information matrix evaluated at $\hat{\btheta}^{\textsc{MLE}}$. \add{The above result implies that \textsc{Waldo}} would enjoy the same asymptotic properties typical of the Wald test, \add{making} it a pivotal test statistic. \add{On the other hand, this does not mean that Wald and \textsc{Waldo} will give the same results for small $n$: indeed, in Section~\ref{sec: waldo_stat_properties} and Appendix~\ref{sec: additional_neural_post}, we demonstrate that \textsc{Waldo} can benefit from a prior over $\btheta$ that is consistent with the data to achieve smaller confidence sets, whereas the Wald test statistic only depends on the likelihood.}

\paragraph{Likelihood-Free Frequentist Inference (LF2I)} \textsc{Waldo} expands on the LF2I framework formalized in \cite{dalmasso2021likelihood}, which \add{proposed a fast construction of Neyman confidence sets using quantile regression to bypass large-sample approximations or expensive Monte-Carlo simulations.} \add{In its original formulation, the LF2I machinery includes three modular procedures which, respectively,}  \textbf{(i)} estimate a likelihood-based test statistic via odds ratios, \textbf{(ii)} estimate critical values $C_{\btheta, \alpha}$ via quantile regression, and \textbf{(iii)} check that the constructed confidence sets achieve the desired coverage level for all $\btheta \in \bTheta$. \add{Each module is based on a independent simulated sample from a high-fidelity simulator $F_{\btheta}$.} \add{\textsc{Waldo} replaces} \textbf{(i)} and instead uses posteriors or predictions to compute $\tau^{\textsc{Waldo}}$ in \eqref{eq: waldo_statistic}. \add{We break down the construction of a confidence set (including diagnostics) in the following steps, as outlined in Figure~\ref{fig:waldo_framework} and Algorithm~\ref{algo:waldo_algo}}: 

\begin{algorithm}[!t]
\caption{Confidence set for $\btheta$ via \textsc{Waldo}}
\label{algo:waldo_algo}
%\algorithmicrequire{ Simulated sets $\mathcal{T}, \mathcal{T}^{\prime}, \mathcal{T}^{\prime\prime}$; observed sample $D$; prediction algorithm; quantile regressor; grid of parameter values $\bTheta_{N_{\text{grid}}}$ at which to perform Neyman inversion; desired coverage level $1-\alpha$.}
\begin{algorithmic}[1]
%\footnotesize
\vspace{2mm}
\State \codecomment{Estimate building blocks of test statistic}
%\vspace{1mm}
\State Simulate $\mathcal{T}=\{ (\btheta^{(j)}, \mathcal{D}^{(j)}) \}_{j=1}^{B}$
\If{prediction algorithm}
\State Estimate $\mathbb{E}[\btheta|\mathcal{D}]$ on $\mathcal{T}$ under squared error loss
\State Compute $\{\mathbf{z}^{(j)}=(\btheta^{(j)} - \mathbb{E}[\btheta|\mathcal{D}^{(j)}])^2\}_{j=1}^B$
\State Estimate $\mathbb{V}[\btheta|\mathcal{D}] = \mathbb{E}[\mathbf{z}|\mathcal{D}]$ under squared error loss
\ElsIf{posterior estimator}
\State Estimate posterior distribution $p(\btheta|\mathcal{D})$ on $\mathcal{T}$
\EndIf
\vspace{2mm}
\State \codecomment{Estimate critical values}
%\vspace{1mm}
\State Simulate $\mathcal{T}^{\prime}=\{ (\btheta^{(j)}, \mathcal{D}^{(j)}) \}_{j=1}^{B^{\prime}}$
\If{prediction algorithm}
\State Predict $\{\hat{\mathbb{E}}[\btheta|\mathcal{D}^{(j)}], \hat{\mathbb{V}}[\btheta|\mathcal{D}^{(j)}]\}_{j=1}^{B^{\prime}}$
\ElsIf{posterior estimator}
\For{each $\mathcal{D} \text{ that appears in } \mathcal{T}^{\prime}$}
\State Draw $N$ samples from $\hat{p}(\btheta|\mathcal{D})$
\State $\hat{\mathbb{E}}[\btheta|\mathcal{D}] \approx \frac{\sum_i \btheta_i}{N}$
\State $\hat{\mathbb{V}}[\btheta|\mathcal{D}] \approx \frac{\sum_i (\btheta_i - \hat{\mathbb{E}}[\btheta|\mathcal{D}])(\btheta_i - \hat{\mathbb{E}}[\btheta|\mathcal{D}])^T}{N-1}$
\EndFor
\EndIf
%\vspace{1mm}
\State Compute $\{\hat{\tau}^{\textsc{Waldo}}(\mathcal{D}^{(j)};\btheta^{(j)})\}_{j=1}^{B^{\prime}}$
\State Estimate critical values $C_{\btheta,\alpha}$ via quantile regression of $\hat{\tau}^{\textsc{Waldo}}(\mathcal{D};\btheta)$ on $\btheta$
\vspace{2mm}
\State \codecomment{Neyman inversion}
%\vspace{0.5mm}
\If{prediction algorithm}
\State Predict $\hat{\mathbb{E}}[\btheta|D]$ and $\hat{\mathbb{V}}[\btheta|D]$
\ElsIf{posterior estimator}
\State Draw $N$ samples from $\hat{p}(\btheta|D)$
\State $\hat{\mathbb{E}}[\btheta|D] \approx \frac{\sum_i \btheta_i}{N}$
\State $\hat{\mathbb{V}}[\btheta|D] \approx \frac{\sum_i (\btheta_i - \hat{\mathbb{E}}[\btheta|D])(\btheta_i - \hat{\mathbb{E}}[\btheta|D])^T}{N-1}$
\EndIf
\State Predict $\hat{C}_{\btheta_0,\alpha} \text{ } \forall \btheta_0 \in \bTheta_{grid}$
%\vspace{0.5mm}
\State Initialize $\mathcal{R}(D) \gets \emptyset$
\For{$\btheta_0 \in \bTheta_{grid}$}
\If{$\hat{\tau}^{\textsc{Waldo}}(D;\btheta_0) \leq \hat{C}_{\btheta_0; \alpha}$}
%\vspace{0.5mm}
\State $\mathcal{R}(D) \gets \mathcal{R}(D) \cup \{\btheta_0\}$
\EndIf
\EndFor
\vspace{2mm}
\State \textbf{return} confidence set $\mathcal{R}(D)$
\end{algorithmic}
\end{algorithm}

\textbf{(i)} \textbf{Estimate the test statistic via prediction algorithms or neural posterior estimators.} Use the simulated set $\mathcal{T}=\{ (\btheta^{(j)}, \mathcal{D}^{(j)}) \}_{j=1}^{B}$, where $\btheta$ can be drawn from any prior distribution $\pi_{\btheta}$, to estimate $\mathbb{E}[\btheta|\mathcal{D}]$ and $\mathbb{V}[\btheta|\mathcal{D}]$. This can be done by choosing between two methods: if using a prediction algorithm, we can leverage the fact that they approximate the conditional mean of the outcome variable given the inputs $\mathcal{D}$, when minimizing the squared error loss (lines 4-6 in Algorithm~\ref{algo:waldo_algo}). Conversely, if using modern neural posterior estimators (such as normalizing flows \citep{papamakarios2021normalizing}), we can approximate $\mathbb{E}[\btheta|\mathcal{D}]$ and $\mathbb{V}[\btheta|\mathcal{D}]$ via Monte Carlo sampling from the estimated posterior distribution (lines 16-18 in Algorithm~\ref{algo:waldo_algo});

\textbf{(ii)} \textbf{Estimate critical values via quantile regression.} Estimate $C_{\btheta,\alpha} \coloneqq F_{\hat{\tau}^{\textsc{Waldo}}}^{-1}(1-\alpha|\btheta)$ by learning the conditional $(1-\alpha)$-quantile of $\hat{\tau}^{\textsc{Waldo}}(\mathcal{D};\btheta)$ using quantile regression over a simulated set $\mathcal{T}^{\prime}=\{ (\btheta^{(j)}, \mathcal{D}^{(j)}) \}_{j=1}^{B^{\prime}}$, where $\btheta$ is drawn uniformly ($r_{\btheta}$ in Figure~\ref{fig:waldo_framework}) over $\bTheta$ to allow calibration $\forall \btheta \in \bTheta$;

\textbf{(i)} $+$ \textbf{(ii)} \textbf{Neyman inversion.} Once $D$ is observed, evaluate $\hat{\tau}^{\textsc{Waldo}}(D;\btheta_0)$ and $\hat{C}_{\btheta_0; \alpha}$ over a fine grid of parameters $\btheta_0 \in \bTheta$, and retain all $\btheta_0$ for which the corresponding test does not reject the null:
\begin{align*}
    \label{eq:confidence_set_neyman}
    \mathcal{R}(D) = \{ \btheta_0 \in \bTheta : \tau^{\textsc{Waldo}}(D;\btheta_0) \leq \hat{C}_{\btheta_0, \alpha}\}.
\end{align*}
As we show in Appendix~\ref{sec: waldo_theory}, step \textit{\textbf{(ii)}} leads to valid level-$\alpha$ hypothesis tests as long as the quantile regressor is well estimated, which then implies that $\mathcal{R}(D)$ satisfies conditional coverage (Eq.~\ref{eq:conditional_coverage}) at level $1-\alpha$, regardless of the true value of $\btheta$ and of the size $n$ of the observed sample $D$.

\textbf{(iii)} \textbf{Coverage diagnostics.} To check that the constructed confidence sets indeed achieve the desired level of conditional coverage, we leverage the diagnostics procedure introduced in \cite{dalmasso2021likelihood}. In detail: simulate a set $\mathcal{T}^{\prime\prime}=\{ (\btheta^{(j)}, \mathcal{D}^{(j)}) \}_{j=1}^{B^{\prime\prime}}$ and construct a confidence region for each  $\mathcal{D}^{(j)} \in \mathcal{T}^{\prime\prime}$. Then model $\mathds{1}\{\btheta^{(j)} \in \mathcal{R}(\mathcal{D}^{(j)})\}$ as a function of $\btheta^{(j)}$ adopting a suitable probabilistic classification method. By definition, this will estimate $\mathbb{E}[\mathds{1}\{\btheta \in \mathcal{R}(\mathcal{D}\}|\btheta] = \mathbb{P}[\btheta \in \mathcal{R}(\mathcal{D})|\btheta]$ across the whole parameter space. \add{Note that this module is completely \textit{independent} from \textbf{(i)} and \textbf{(ii)}. As such, it can be used to to check the empirical conditional coverage of any uncertainty estimate, as illustrated in Section~\ref{sec: comput_properties} for Neyman confidence sets where critical values are estimated via Monte Carlo sampling, in Section~\ref{sec: mixture_bayesianLFI} for posterior credible regions, and in Section~\ref{sec: muons} for prediction sets from the output of a CNN.}

\subsection{Statistical Properties: Coverage and Power}
\label{sec: waldo_stat_properties}

We now show that the coverage guarantees of \textsc{Waldo} are independent from the prior distribution, which can also be chosen to increase power. We do so through univariate Gaussian examples with analytically computable solutions. Since $p=1$, we use simple prediction algorithms to estimate $\mathbb{E}[\btheta|\mathcal{D}]$ and $\mathbb{V}[\btheta|\mathcal{D}]$. See Appendix~\ref{sec: details_stat_properties} for details.

%\textsc{Waldo} guarantees that the constructed confidence sets are always conditionally valid (Equation~\ref{eq:conditional_coverage}) \textit{and} can leverage prior information. We illustrate these statistical properties through

\label{sec: coverage}
\begin{figure}[!b]
    %\floatbox[{\capbeside\thisfloatsetup{capbesideposition={right,top}}}]{figure}[\FBwidth]
    {\caption{
    \textbf{\textsc{Property I:} \textsc{Waldo} guarantees conditional coverage across $\bTheta$, regardless of the specified prior.} 
    \add{Prior: $\theta \sim \mathcal{N}(0, 2)$. Likelihood: $\mathcal{D}|\theta \sim \mathcal{N}(\theta, 1)$}. \textit{Left:} median of upper/lower bounds of constructed parameter regions. \textit{Right:} empirical coverage computed numerically using 100,000 samples for each $\theta$ over a fine grid in~$\Theta$ \add{(i.e., not using coverage diagnostics)}.}
    \label{fig:coverage_toy}}
    {\includegraphics[width=1 \textwidth]{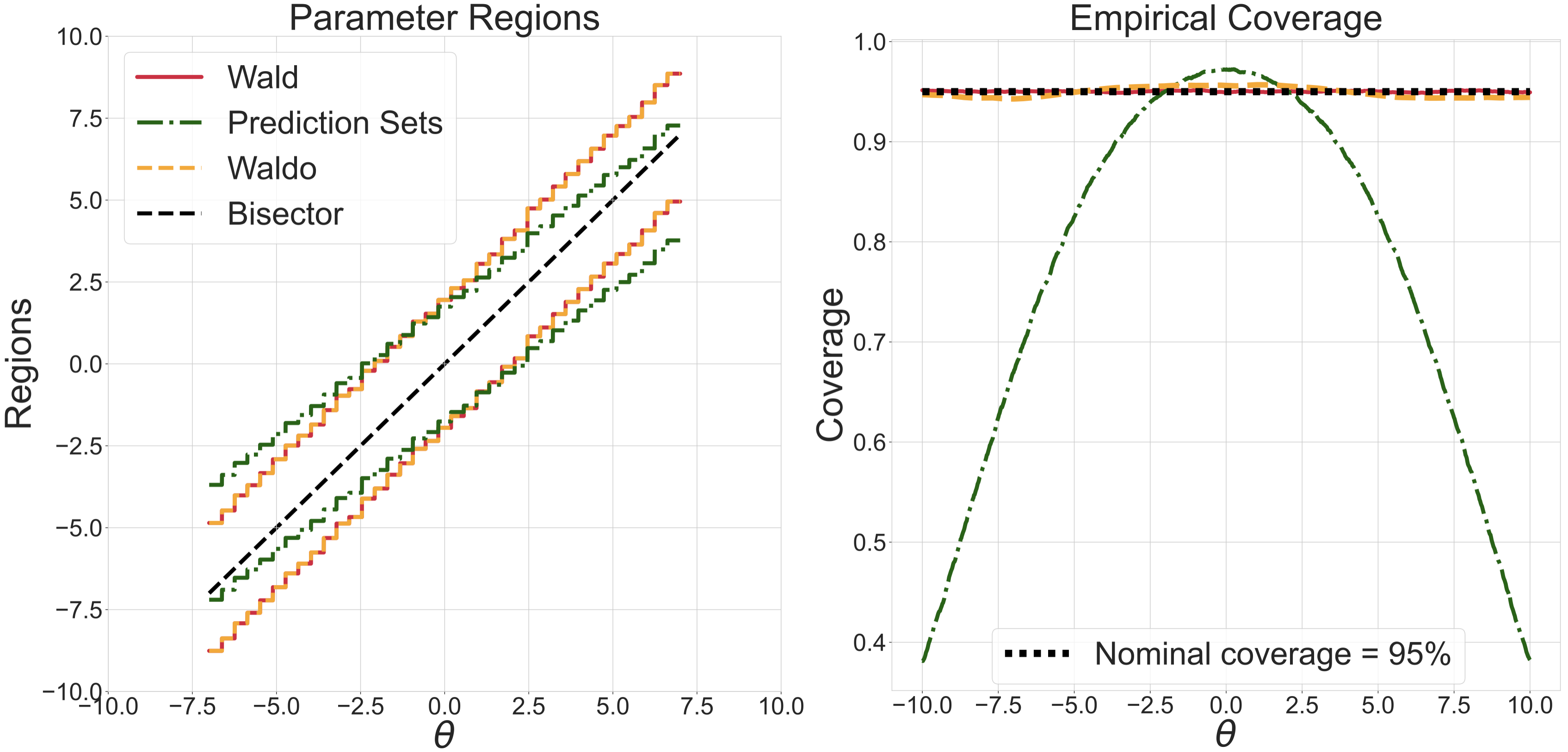}}
\end{figure}

\paragraph{\textsc{Property I:} \textsc{Waldo} guarantees conditional coverage across $\bTheta$, regardless of the specified prior.} Scientists sometimes have domain-specific knowledge that can guide inference through the elicitation of a prior distribution over the parameters of interest. \add{The goal is to introduce a bias to help quantifying the uncertainty, but if the prior happens to be at odds with the data, then this bias can be harmful and cause posteriors to be overconfident and smaller
%\add{tighter} 
than they should be \citep{hermans2021averting}.} Ideally, we would want the \add{coverage} guarantees of any estimated parameter region to be preserved under this bias. \add{In this example, we assume $\theta \sim \mathcal{N}(0, 2), \text{ } \mathcal{D}|\theta \sim \mathcal{N}(\theta, 1)$. As Figure~\ref{fig:coverage_toy} shows, confidence sets for $\theta$ (left panel) constructed through Neyman inversion of a series of Wald tests guarantee the correct conditional coverage (right panel), since Wald is only influenced by the likelihood. Conversely, prediction sets ($\mathbb{E}[\theta|\mathcal{D}] \pm 1.96\sqrt{\mathbb{V}[\theta|\mathcal{D}]}$) are influenced by the prior through the bias induced in the point predictions, which increases with the distance from the prior mean and results in strong under-coverage. \textsc{Waldo} exploits the same inputs of prediction sets ($\mathbb{E}[\btheta|\mathcal{D}]$ and $\mathbb{V}[\btheta|\mathcal{D}]$), but corrects this problem by calibrating the critical values via quantile regression, hence guaranteeing conditional coverage. Note that we only use a single observation ($n=1$) for each confidence set.}
    
\begin{figure}[!t]
    %\floatbox[{\capbeside\thisfloatsetup{capbesideposition={right,top}}}]{figure}[\FBwidth]
    {\caption{\textbf{\textsc{Property II:} \textsc{Waldo} exploits prior information and achieves higher power.} Power curves computed by recording the \add{number of times a wrong value of $\theta$ is correctly outside the confidence set over 1{,}000 repetitions. Likelihood: $\mathcal{D} \sim \mathcal{N}(40, 1)$. \textit{Left:} Wald and \textsc{Waldo} are equivalent when $\theta \sim \mathcal{U}(35, 45)$ . \textit{Right:} \textsc{Waldo} has higher power when $\theta \sim \mathcal{N}(40, 1)$.}}
    \label{fig:power_gaussian}}
    {\includegraphics[width=1 \textwidth]{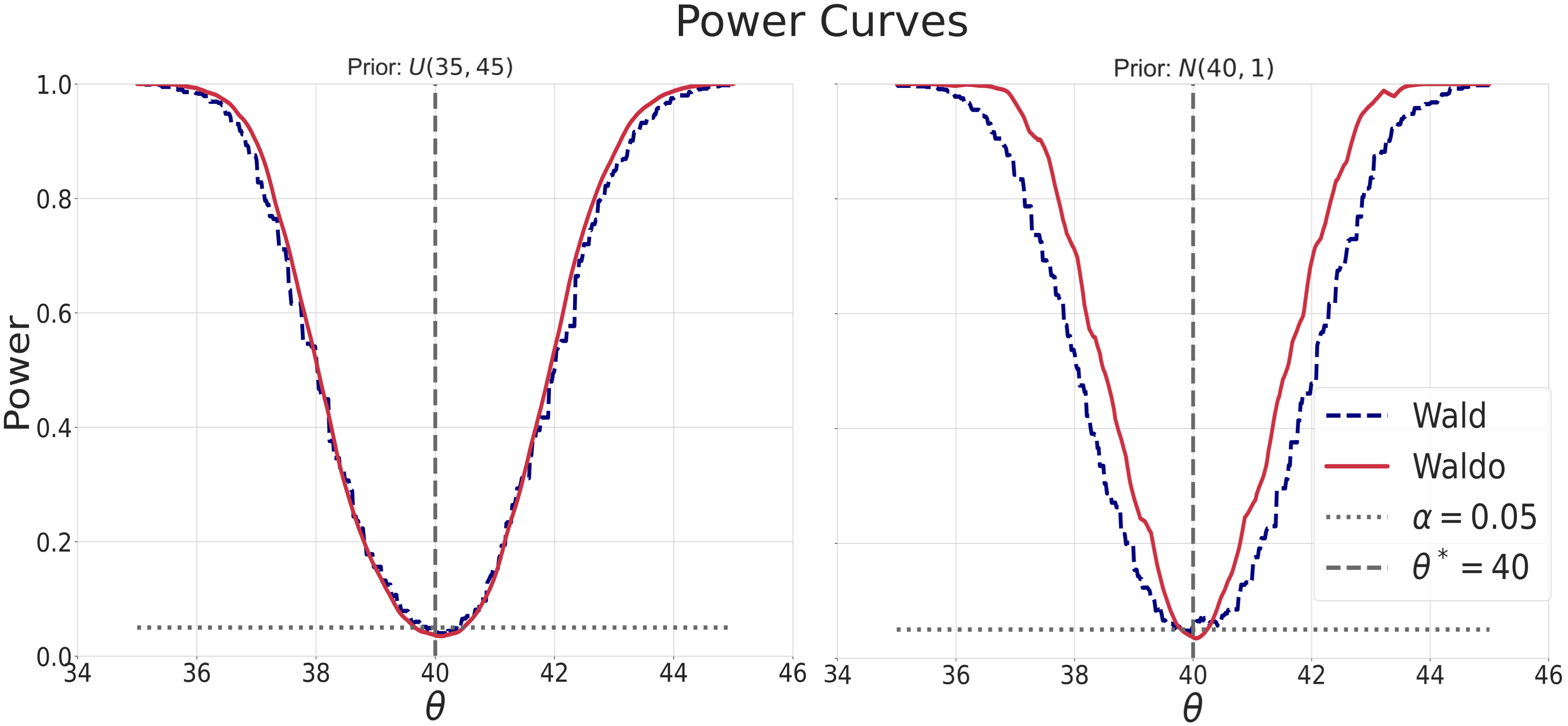}}
\end{figure}

\paragraph{\textsc{Property II:} \textsc{Waldo} exploits prior information and achieves higher statistical power.} When the prior is correctly specified, we would like to \add{leverage the induced bias to increase the power of the inverted tests and produce tighter constraints on the parameters, while} retain\add{ing} conditional coverage. \add{Here we simulate data from a unique ``true'' Gaussian likelihood $\mathcal{D}|\theta \sim \mathcal{N}(\theta=40, 1)$, and investigate the effect that the informativeness of the prior has on the power of the resulting tests.} \add{As Figure~\ref{fig:power_gaussian} shows,
\textsc{Waldo} and Wald coincide when the prior is uninformative ($\theta \sim \mathcal{U}(35, 45)$; left panel), but the former has higher power when the prior is instead correctly specified ($\theta \sim \mathcal{N}(40, 1)$; right panel), thereby leading to smaller confidence sets.}

\subsection{Computational Properties}
\label{sec: comput_properties}

\begin{figure}[!b]
    %\floatbox[{\capbeside\thisfloatsetup{capbesideposition={right,top},capbesidewidth=0.15\textwidth}}]{figure}[\FBwidth]
    {\caption{\small
    \textbf{Quantile regression (QR) is orders of magnitude more efficient than Monte Carlo (MC) in terms of the number of simulations $B^\prime$ required to achieve correct coverage.} Each panel shows the fraction of samples (out of 1{,}000 total) for which the selected method to estimate critical values achieves approximately correct coverage ($\mathbb{P}(\btheta \in \mathcal{R}(\mathcal{D})|\btheta) \in [0.95 \pm 0.03]$). Prior: $\theta \sim \mathcal{N}(0, 0.1\cdot\mathbf{I})$. Likelihood: $\mathcal{D}|\theta \sim \mathcal{N}(\btheta, 0.1\cdot\mathbf{I})$. In both cases, we used normalizing flows to estimate the posterior.}
    \label{fig:high_dim_gauss}}
    {\includegraphics[width=1\textwidth]{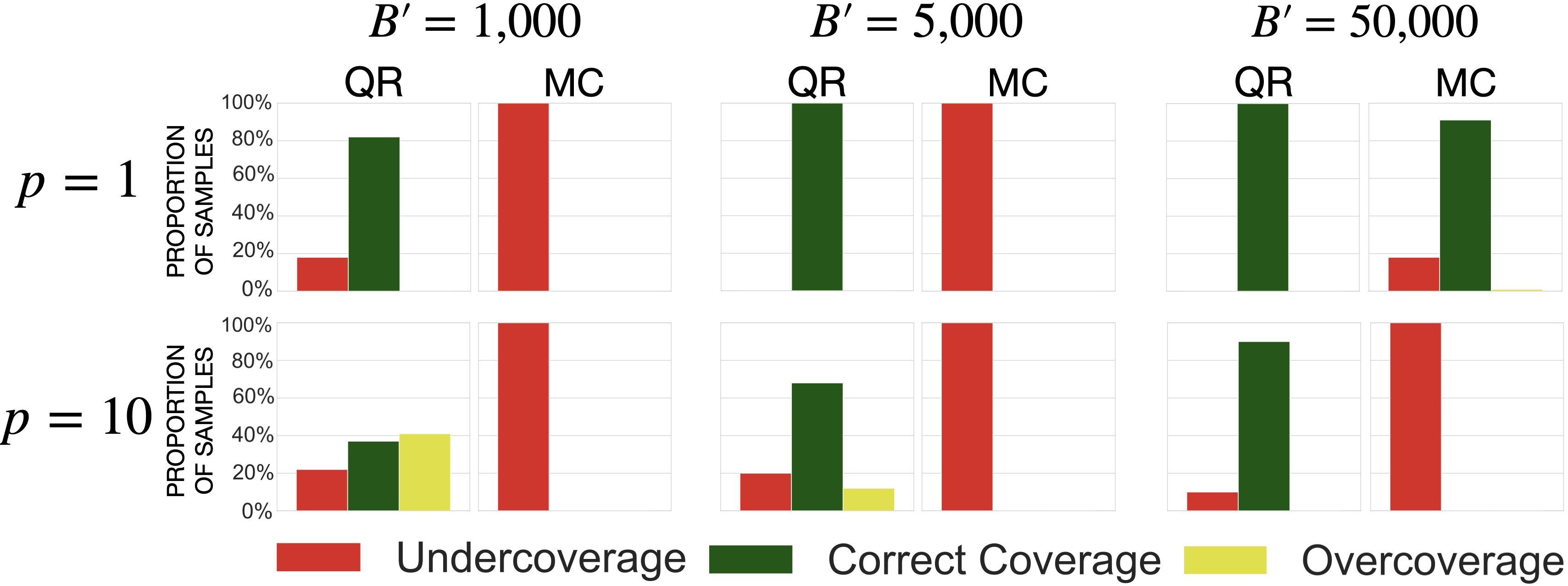}}
\end{figure}

\add{\paragraph{Scaling with high-dimensional parameters}As mentioned in Section~\ref{sec: waldo_methodology}, \textsc{Waldo} exploits a simulated set sampled uniformly\footnote{Technically, we only need to sample from a distribution that places mass on all $\bTheta$.} over $\bTheta$ to estimate critical values via quantile regression and guarantee coverage across the whole parameter space. While this might seem a daunting requirement, the only alternative to guarantee conditional coverage is to resort to Monte Carlo approaches that sample many times \textit{at each} $\btheta \in \bTheta$. As Figure~\ref{fig:high_dim_gauss} shows, \textsc{Waldo} requires several orders of magnitude less simulations to achieve the correct calibration. This is true already when $p=1$, and is even more evident when $p=10$.}

\add{\paragraph{Quality of models} \textsc{Waldo} relies on two estimation procedures (\textbf{(i)} and \textbf{(ii)} below) to construct the confidence set itself. The accuracy of the results relies on the estimation quality of these models and on the number of simulations $B$ and $B^\prime$ that are available. In addition, there is a diagnostics procedure \textbf{(iii)} to estimate the conditional coverage of the final confidence sets, as a separate check that Equation~\ref{eq:conditional_coverage} indeed holds. 

\textbf{(i) Test statistic.} The quality of prediction algorithms and posterior estimators is positively correlated with the power of the resulting tests. As the precision in the estimates of $\mathbb{E}[\btheta|\mathcal{D}]$ and $\mathbb{V}[\btheta|\mathcal{D}]$ decreases, the variance of the test statistics increases, which implies more conservative critical values and larger confidence regions. A good prior distribution will clearly help in achieving more precise estimates in regions of interest in the parameter space. 

\textbf{(ii) Critical values.} As we prove in Appendix~\ref{sec: waldo_theory}, conditional coverage is achieved as long as the quantile regressor is well estimated. In practice, we observe that little hyper-parameter optimization is needed and that the number of simulations required to achieve well-calibrated critical values is usually a small fraction of those needed for the test statistic. 

\textbf{(iii) Diagnostics.} The quality of the probabilistic classifier used to check the empirical coverage probability affects only the reliability of the diagnostics. Note that this module is completely independent of the others, and we can check its quality by inspecting the cross-entropy loss, and the standard errors and confidence bands on the estimates that common statistical packages provide (e.g., \texttt{MGCV} in \texttt{R}).}

\section{RESULTS}
\label{sec: experiments}

We assess the performance of \textsc{Waldo} on two challenging experiments. In the first example (Section~\ref{sec: mixture_bayesianLFI}), we show how to use a posterior distribution estimated via normalizing flows to compute valid confidence regions, and how prior information can improve precision.
The second example (Section~\ref{sec: muons}) tackles a complex particle energy reconstruction problem in high-energy physics: we leverage predictions from a custom CNN to construct confidence intervals with correct coverage and high power.

\subsection{Confidence Sets from Neural Posteriors}
\label{sec: mixture_bayesianLFI}

%\tommaso{This is really too small-can we increase the size of this figure? Do we have space limitations?}\ann{@Luca: see my suggestion in comments}
\begin{figure*}[!t]
    %\floatbox[{\capbeside\thisfloatsetup{capbesideposition={left,top}}}]{figure}[\FBwidth]
    {\caption{\small
    \textbf{\textsc{Waldo} converts posterior distributions into confidence regions with correct conditional coverage and high power}. \textit{Left Panel - Top:} Examples of 95\% credible regions (\textcolor{Blue}{blue}) from posteriors estimated with normalizing flows and a Gaussian $\mathcal{N}(\mathbf{0}, 2\mathbf{I})$ prior (\textcolor{Gray}{gray}) for different values of the true unknown parameter $\boldsymbol{\theta}^*$ (\textcolor{red}{red star}). \textit{Right Panel - Top:} Credible regions have conditional coverage close to the nominal level only in a neighborhood of the prior, and severely undercover everywhere else. \textit{Left Panel - Bottom:} Corresponding 95\% \textsc{Waldo} confidence sets (\textcolor{OliveGreen}{green}), derived from the same posterior estimates used for the top row. \textit{Right Panel - Bottom:} Conditional coverage for \textsc{Waldo} confidence sets achieves the nominal 1-$\alpha$ level everywhere, where $\alpha=0.05$.
    }
    \label{fig:waldo_bayesian_lfi}}
    {\includegraphics[width=0.8\textwidth]{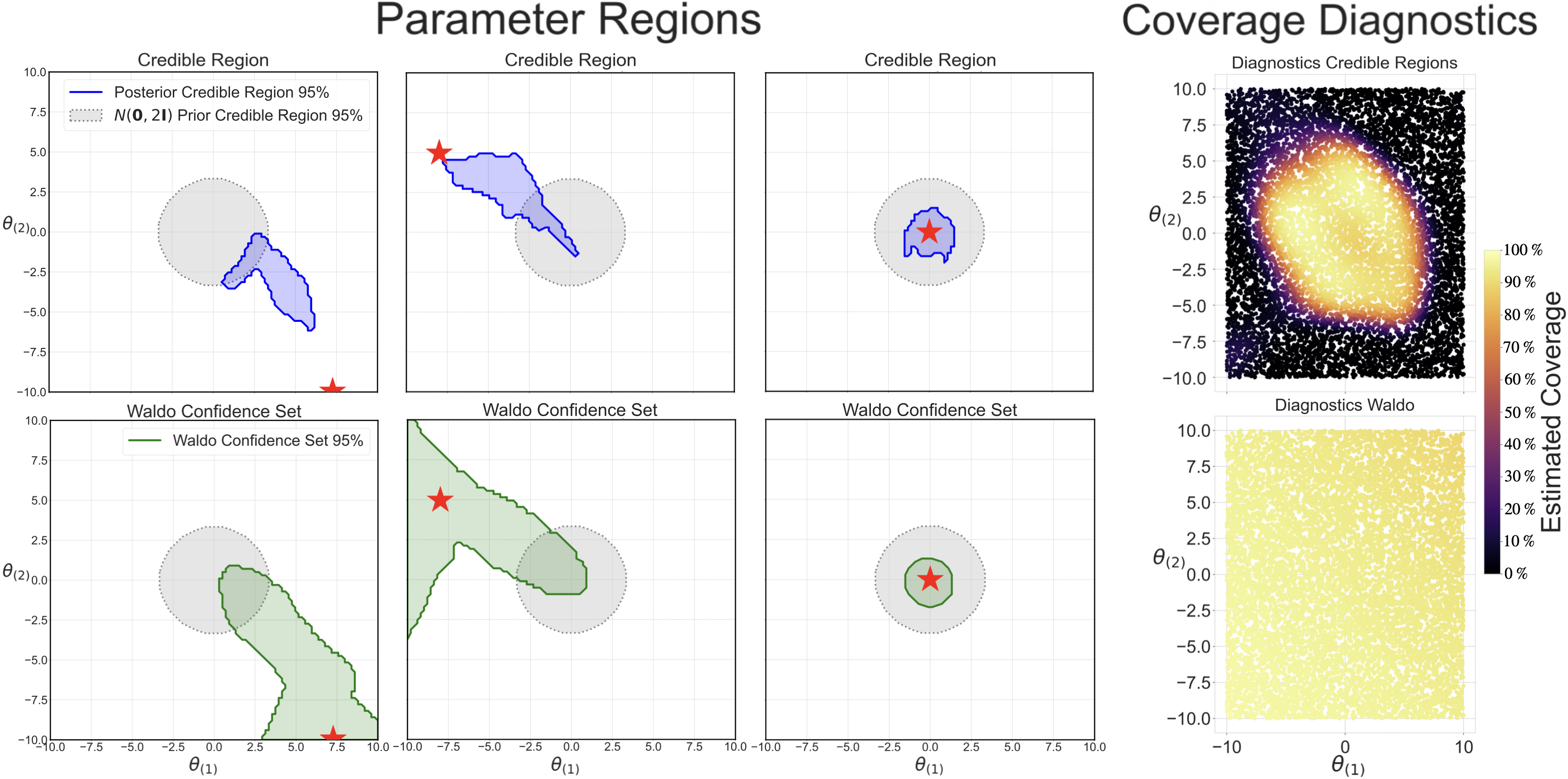}}
\end{figure*}

This inference task was introduced in \cite{sisson2007sequential} and has become a standard benchmark in the SBI literature \citep{clarte2021componentwise, toni2009approximate, simola2021adaptive, lueckmann2021benchmarking}. It consists of estimating the (common) mean of the components of a two-dimensional Gaussian mixture, with one component having much broader covariance:
$\mathcal{D}|\btheta \sim \frac{1}{2}\mathcal{N}(\btheta, \mathbf{I}) + \frac{1}{2}\mathcal{N}(\btheta, 0.01\cdot\mathbf{I}),$
where $\btheta \in \mathbb{R}^2$ and $n=1$\footnote{\textsc{Waldo} works for an observed sample of any size, but we had to use $n=1$ because the \texttt{SBI} Python library we used to estimate the posterior does not yet support larger sample sizes for NPE.}. We estimate $p(\btheta|\mathcal{D})$ using the implementation of Neural Posterior Estimators (NPE) of \cite{nflows} through the \texttt{SBI} software package \citep{tejero-cantero2020sbi}, and report results obtained with two different priors: $\btheta \sim \mathcal{N}(\mathbf{0}, 2\cdot\mathbf{I})$ and $\btheta \sim \mathcal{U}([-10, 10]^2)$ (the latter in Appendix~\ref{sec: additional_neural_post}). We estimate the critical values with a 2-layer neural network minimizing the quantile loss. Simulated datasets used for training are of the following sizes: $B=100{,}000$, \add{$B^\prime=30{,}000$ when using a Gaussian prior}. Conditional mean and variance were approximated with $50{,}000$ Monte Carlo samples from the neural posterior.

\add{The first four panels on the left of Figure~\ref{fig:waldo_bayesian_lfi} show examples of 95\% credible regions (top) and \textsc{Waldo} confidence sets (bottom) obtained from the same posterior distribution, when the true parameter is far from the prior. If the data is at odds with the prior, then the induced bias leads to credible regions that severely undercover across the parameter space, as it is shown at the top of the rightmost panel, where the coverage probability for credible regions reaches values as low as 0-10\%. \textsc{Waldo} can correct for this bias and output larger confidence sets which account for the added uncertainty, thereby leading to correct conditional coverage everywhere (bottom of rightmost panel). This is the same behaviour seen in the first example of Section~\ref{sec: waldo_stat_properties}, although for a more complex setting and for a posterior estimator.

Conversely, when the prior is consistent with the data (Figure~\ref{fig:waldo_bayesian_lfi}, right two panels of “Parameter Regions”), \textsc{Waldo} is not overly conservative and leverages the additional information to tighten the constraints on the parameters, closely tracking the size of the posterior credible region. In Appendix~\ref{sec: additional_neural_post}, we also show that, over many independent observations, the average size of \textsc{Waldo} confidence sets is indeed smaller when using an informative prior than when using a Uniform over $\bTheta$. These results closely mimic those seen in the second example of Section~\ref{sec: waldo_stat_properties}.
}

\subsection{Confidence Sets for Muon Energies using CNN Predictions}
\label{sec: muons}

\begin{figure*}[!t]
    % \floatbox[{\capbeside\thisfloatsetup{capbesideposition={right,top},capbesidewidth=0.15\textwidth}}]{figure}[\FBwidth]
    {\caption{\small 
    \textbf{\textsc{Waldo} guarantees the nominal coverage level, and yields smaller confidence intervals (more precise estimates of muon energy) with the higher-granularity (“full”) calorimeter data.} \textit{Left:} Energy deposited by a $\theta \approx 3.2$ TeV muon entering a calorimeter with $32\times32\times50$ cells. \textit{Center:} \textsc{Waldo} (\textcolor{Blue}{blue}, \textcolor{Dandelion}{orange}, \textcolor{Maroon}{red} in the right two panels) guarantees nominal coverage ($68.3\%$), while $1\sigma$ prediction intervals (\textcolor{OliveGreen}{green}) under- or over-cover in different regions of $\Theta$. \textit{Right:} Median lengths of constructed intervals: shorter intervals imply higher precision in the estimates. Prediction sets are on average wider than the corresponding confidence sets, using the same data.}
    \label{fig:muons68}}
    {\includegraphics[width=1\textwidth]{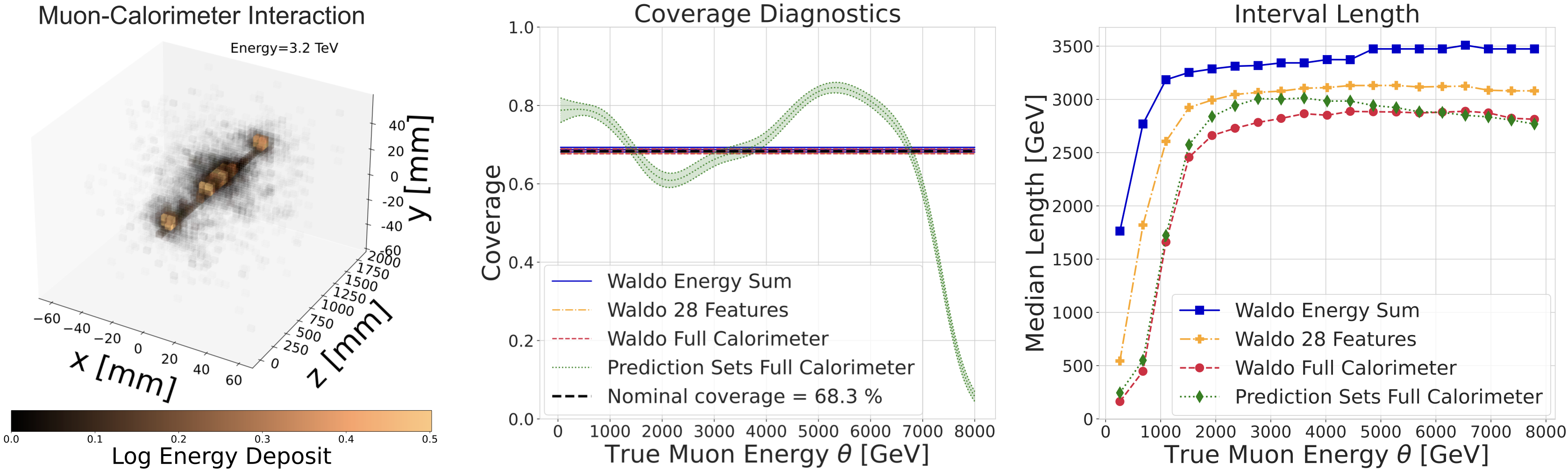}}
\end{figure*}

We now discuss the performance of \textsc{Waldo} on an application of interest to fundamental research: estimating the energy of muons at a future particle collider. Muons are a heavier replica of electrons; they are produced in sub-nuclear reactions involving electroweak interactions. Muons are also excellent probes of new phenomena: in fact, their detection and measurement has been key to several crucial discoveries in the past decades\add{, including the Higgs boson} \citep{augustin1974discovery, herb1977observation, cdf1995observation, aad2012observation, chatrchyan2012observation}. \add{Traditionally,} the energy of a muon is determined from the curvature of its trajectory in a magnetic field, but at energies above a few TeV this methods breaks down as trajectories become indistinguishable from straight paths even within the strongest practically achievable fields. Searching for viable alternatives, it has been observed \citep{kieseler2022calorimetric, dorigo2022deep} that both the pattern and the magnitude of small radiative energy losses that muons withstand in traversing dense and finely segmented calorimeters can be used to infer the incident muon energy, \add{leveraging the capacity of modern deep learning architectures. Nonetheless, the above work also clearly showed that predictions of $\theta$ suffered from a strong bias, mainly due to the high nonlinearity of the response at very high energies.
%low signal-to-noise ratio in the calorimeter data at very high energies 
%\tommaso{Hmm, not sure I would say that. The bias is mainly due to the reduction of resolution (observables grow logarithmically) as E becomes very high. Can we rewrite as "mainly due to non-linearity of response at very high energy." ?} 
Motivated by this problem, we pose two questions: \textbf{(i)} Can we construct confidence sets with correct coverage of the true energy of muons using the information contained in the pattern and magnitude of radiative deposits in a dense calorimeter? \textbf{(ii)} Is it possible to extract additional information from finer segmentations of the calorimeter to allow for tighter constraints (i.e., smaller confidence sets with correct coverage) on muon energy estimates? Quantifying the latter would allow scientists to optimize their detector designs, since manufacturing very small calorimeter cells is expensive.}

We have available $886{,}716$ 3D input “images” $\mathbf{x}$ and scalar muon energies $\theta$ obtained through \textsc{Geant4} \citep{agostinelli2003geant4}, a high-fidelity stochastic simulator. \add{See Figure~\ref{fig:muons68} (left panel) for an illustration of one simulated $\mathbf{x}_i$ for a particular $\theta_i$.} The data are available in \cite{kieseler_jan_2021_5163817}. As the interest is on constraining muon energies as much as possible while guaranteeing conditional coverage, we use three versions of the same dataset with increasing dimensionality: a 1D input equal to the sum over all calorimeter cells with deposited energy $E > 0.1$ GeV, for each muon; 28 custom features extracted from the spatial and energy information of the calorimeter cells (see \cite{kieseler2022calorimetric}); and the full calorimeter measurements ($\mathbf{x}_i \in \mathbb{R}^{51,200})$. For the first two datasets, we estimate $\mathbb{E}[\btheta|\mathcal{D}]$ and $\mathbb{V}[\btheta|\mathcal{D}]$ via Gradient Boosting \citep{chen2016xgboost}. For the full calorimeter data, we rely on the CNN developed by \cite{kieseler2022calorimetric}. We use Gradient Boosting for quantile regression \citep{pedregosa2011scikit}.

\add{Answering \textbf{(i)} \add{affirmatively}, Figure~\ref{fig:muons68} (center) shows that confidence sets constructed with \textsc{Waldo} achieve exact conditional coverage ($68.3\%$) regardless of the dataset used. The corresponding $1\sigma$ prediction intervals ($\mathbb{E}[\btheta|\mathcal{D}] \pm \sqrt{\mathbb{V}[\btheta|\mathcal{D}]}$) using full calorimeter data, instead, exhibit over- or under-coverage in different regions over $\bTheta$, which in the latter case means that prediction sets contain the true value with much lower probability than anticipated. As for question~\textbf{(ii)}, we make two observations (see Figure~\ref{fig:muons68}; right panel): First, using the raw higher-dimensional energy deposits \add{with \textsc{Waldo}} allows to reduce the uncertainty around muon energies. Second, confidence sets constructed with \textsc{Waldo} are even shorter than the corresponding prediction intervals, while also guaranteeing conditional coverage.}

\section{DISCUSSION}
\label{sec: discussion}

We presented \textsc{Waldo}, a novel method to construct confidence sets with correct finite-$n$ conditional coverage by leveraging prediction algorithms and posterior estimators for inverse problems. \textsc{Waldo} relies on a regression-based Neyman construction, which requires orders of magnitude fewer simulations than traditional Monte Carlo approaches to be well calibrated across the parameter space (see Section~\ref{sec: comput_properties}). Nonetheless, our method still needs a simulator that is both high-fidelity -- to draw inferences that reflect the true data-generating process -- and fast -- to simulate sufficiently large training sets to accurately learn the key quantities of \textsc{Waldo}: the test statistics, the critical values, and the coverage diagnostics, as discussed in Section~\ref{sec: comput_properties}. \textsc{Waldo} disentangles the \textit{coverage} guarantees of the confidence region from the choice of the prior distribution.
%\ann{Alternatively we could say: ``\textsc{Waldo} disentangles the coverage of the confidence set from the choice of prior, and guarantees {\em validity} regardless of the prior and the number of observations $n$.''}  
To increase 
%\ann{could put ``power'' in italics if we do the same with ``validity'' in the previous sentence} 
\textit{power}, one may be able to leverage domain-specific knowledge (see Sections~\ref{sec: waldo_stat_properties} and ~\ref{sec: mixture_bayesianLFI}), or take advantage of the internal structure of the simulator \citep{Brehmer2020}, with the guarantee that the confidence sets always contain the true parameter with the desired proability. One could also adaptively simulate more data in specific regions of interest in the parameter space. Active learning strategies, and a more formal treatment of the relation between power and priors, are promising areas for future studies. 

Domain sciences, especially the physical sciences, routinely seek to constrain parameters of interest using both theoretical (or simulation) models and experimental data. \textsc{Waldo} provides reliable constraints that can be used to deduce trustworthy scientific conclusions when other uncertainty quantification methods are either unavailable, unreliable or inefficient.

\subsubsection*{Acknowledgments}
\input{acknowledgements}

\bibliographystyle{plainnat}
\bibliography{references}

\appendix
\onecolumn
\input{supplementary}

\end{document}

%% file: acknowledgements.tex
We thank Niccolò Dalmasso for early feedback and discussions on this work, and for providing code previously written for LF2I. We are also indebted to Jan Kieseler and to Giles C. Strong for providing the muon energy data and the structure of the deep neural network employed for the studies described in Section~\ref{sec: muons}, respectively. We also thank Michael Stanley for many valuable discussions on the details of \textsc{Waldo}. This work is supported in part by NSF DMS-2053804, NSF PHY-2020295, and the C3.ai Digital Transformation Institute. RI is grateful for the financial support of CNPq (309607/2020-5 and 422705/2021-7) and FAPESP (2019/11321-9). We are also grateful to Microsoft for providing Azure computing resources for this work.

%% file: supplementary.tex
\section{THEORETICAL RESULTS}
\label{sec: waldo_theory}

We assume that the quantile regression estimator described in Section~\ref{sec: methodology} is consistent
in the following sense:

\begin{Assumption}[Uniform consistency]
\label{assum:quantile_consistent_simple_null} 
Let $ F(\cdot|\btheta)$ be the cumulative  distribution function of the test statistic $\tau(\mathcal{D};\btheta_0)$ conditional on $\btheta$, where $\mathcal{D} \sim F_{\btheta}$. Let $\hat{F}_{B'}(\cdot|\btheta)$ be the estimated conditional distribution function, implied by a quantile regression with a sample $\mathcal{T}^\prime$ of $B^\prime$ simulations $\mathcal{D} \sim F_{\btheta}$.
Assume that the quantile regression estimator is such that
$$\sup_{\tau \in \mathbb{R}}|\hat F_{B^\prime}(\tau|\btheta_0)-  F(\tau|\btheta_0)|\xrightarrow[B^\prime \longrightarrow\infty]{\enskip \mathbb{P} \enskip} 0.$$
\end{Assumption}

Assumption~\ref{assum:quantile_consistent_simple_null} holds, for instance, for quantile regression forests \citep{meinshausen2006quantile}. Next, we show that step \textbf{(ii)} in Section~\ref{sec: waldo_methodology} yields a valid hypothesis test as $B^\prime \rightarrow \infty$. 

\begin{thm}
 \label{thm:valid_tests}
Let $C_{B^\prime} \in \mathbb{R}$ be the 
critical value of the test based on a strictly continuous statistic  $\tau(\mathcal{D};\btheta_0)$ chosen according to step \textbf{(ii)}
for a fixed $\alpha \in (0,1)$. If the quantile estimator satisfies Assumption~\ref{assum:quantile_consistent_simple_null},
then,
$$ \mathbb{P}_{\mathcal{D}|\btheta_0,C_{B^\prime}}(\tau(\mathcal{D};\btheta_0) \geq C_{B^\prime})  \xrightarrow[B^\prime \longrightarrow\infty]{\enskip a.s. \enskip}   \alpha,$$
where $\mathbb{P}_{\mathcal{D}|\btheta_0,C_{B^\prime}}$ denotes the probability integrated over $\mathcal{D}\sim F_{\btheta_0}$ and conditional on the random variable $C_{B^\prime}$.
\end{thm}

If the convergence rate of the quantile regression estimator is known (Assumption \ref{assum:quantile_regression_rate}),  Theorem \ref{thm:valid_tests_rate} provides a finite-$B^\prime$ guarantee on how far the Type-I error of the test will be from the nominal level.

\begin{Assumption}[Convergence rate of the quantile regression estimator]
\label{assum:quantile_regression_rate}
Using the notation of Assumption \ref{assum:quantile_consistent_simple_null}, assume that the quantile regression estimator is such that
$$\sup_{\tau \in \mathbb{R}}|\hat F_{B^\prime}(\tau|\btheta_0)-  F(\tau|\btheta_0)|= \mathcal{O}_p\left(\left(\frac{1}{B^\prime}\right)^{r}\right)$$
for some $r>0$.
\end{Assumption}

\begin{thm}
 \label{thm:valid_tests_rate}
With the notation and assumptions of Theorem \ref{thm:valid_tests}, and if  Assumption~\ref{assum:quantile_regression_rate} also holds,
then,
$$ |\mathbb{P}_{\mathcal{D}|\btheta_0,C_{B^\prime}}(\tau(\mathcal{D};\btheta_0) \geq C_{B^\prime}) - \alpha| = \mathcal{O}_p\left(\left(\frac{1}{B^\prime}\right)^{r}\right).$$
\end{thm}

Proofs of these results can be found in \cite{dalmasso2021likelihood}.

\section{ADDITIONAL EXPERIMENTS}

\subsection{\textsc{Property III:} Estimating the Conditional Variance Matters}
\label{sec: property3}

\begin{figure}[t!]
    %\floatbox[{\capbeside\thisfloatsetup{capbesideposition={right,top}}}]{figure}[\FBwidth]
    {\caption{\textbf{\textsc{Property III:} Estimating the conditional variance matters.} \textit{Left:} Power curves at 95\% confidence level when the true Pareto shape $\theta^*=5$, implying a very skewed data distribution. \textit{Right:} Test statistics and critical values as a function of $\theta$. ($n=10$).}
    \label{fig:power_pareto}}
    {\includegraphics[width=0.6\textwidth]{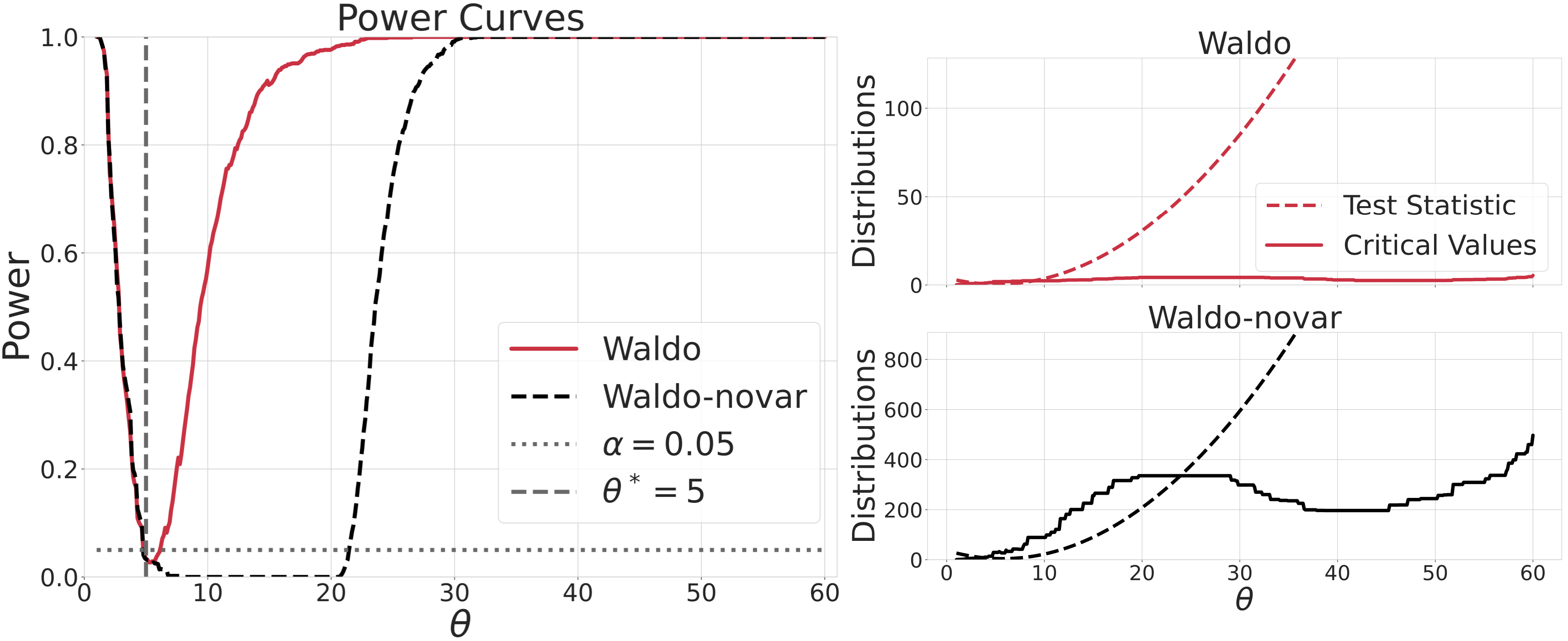}}
\end{figure}

We complete the exposition of the statistical properties of \textsc{Waldo} (Section~\ref{sec: waldo_stat_properties}) by demonstrating the importance of estimating the conditional variance in the test statistic $\tau^{\textsc{Waldo}}$. Recall that in principle any test statistic defined in an LFI setting could be used for our framework. One could then define a simpler “unstandardized” test statistic $\tau^{\textsc{Waldo-novar}}(\mathcal{D};\boldsymbol{\btheta}_0) = (\mathbb{E}[\btheta|\mathcal{D}] - \btheta_0)^T(\mathbb{E}[\btheta|\mathcal{D}] - \btheta_0)$ which does not require estimation of $\mathbb{V}[\btheta|\mathcal{D}]$. It turns out that estimating $\mathbb{V}[\btheta|\mathcal{D}]$ and using $\tau^{\textsc{Waldo}}$ is actually of crucial importance, as it leads to confidence regions of smaller or equal expected size, especially in settings where the conditional variance varies significantly as a function of $\btheta$. Consider, for example, the problem of estimating the shape of a Pareto distribution with fixed scale $x_\mathrm{min}=1$ and true unknown shape $\theta^*=5$, which yields a strongly right-skewed data distribution. Figure~\ref{fig:power_pareto} shows that $\tau^{\textsc{Waldo}}$ has much higher power than $\tau^{\textsc{Waldo-novar}}$ for inferring $\theta$. Dividing by the conditional variance effectively stabilizes the test statistic and makes its distribution over $\mathcal{D}$ pivotal, i.e., independent of $\theta$. This implies that the critical values will be relatively constant over $\theta$ (see top right panel for \textsc{Waldo}), which yields tighter parameter regions due to the curvature of the test statistic.

\subsection{Confidence Sets from Neural Posteriors: Two-Dimensional Gaussian Mixture}
\label{sec: additional_neural_post}

\begin{figure}[!b]
    %\floatbox[{\capbeside\thisfloatsetup{capbesideposition={left,top}}}]{figure}[\FBwidth]
    {\caption{\small \textbf{a) When the prior is uninformative, \textsc{Waldo} can still correct for possible approximation errors in the estimated posterior. b)-c) When the prior is consistent with the data, \textsc{Waldo} tightens the confidence sets, improving the precision with respect to the case using a Uniform prior.} \textit{a)-b):} Posterior credible regions and \textsc{Waldo} confidence sets using different priors. \textit{Right:} Average area of credible regions and \textsc{Waldo} confidence sets across 100 independent samples, reported as the percentage of points retained among those in the evaluation grid.}
    \label{fig:sets_and_area}}
    {\includegraphics[width=0.9\textwidth]{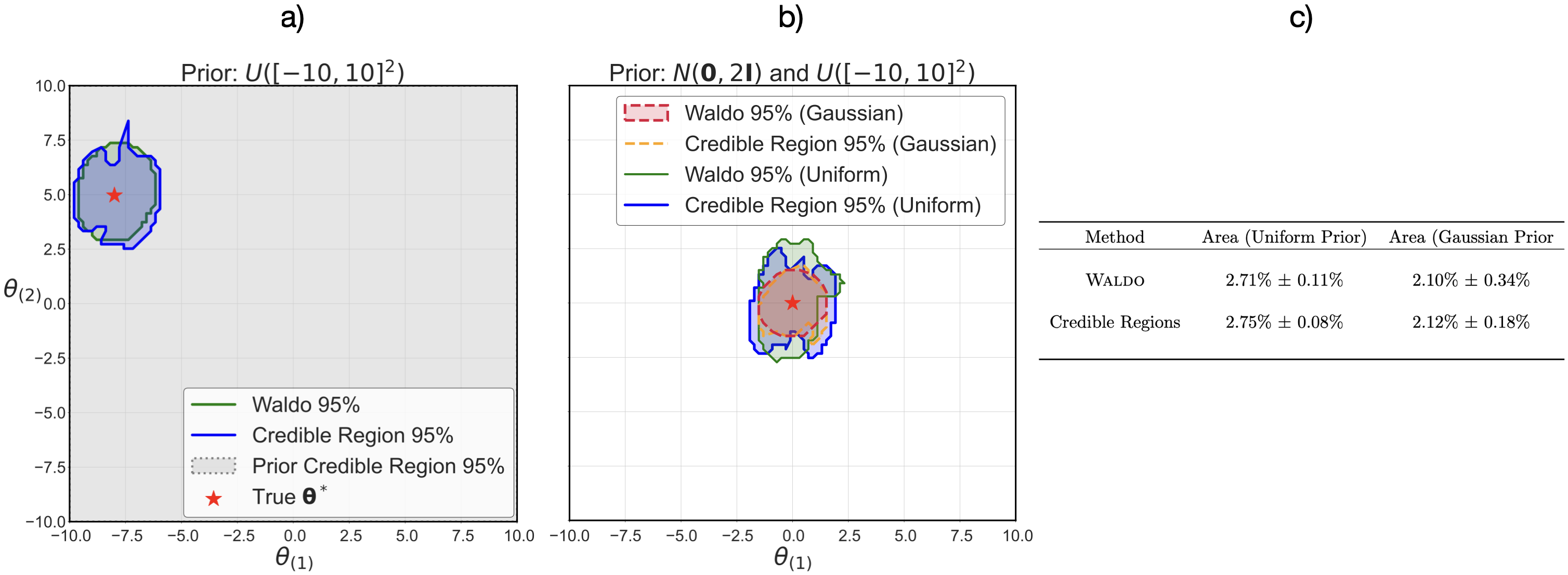}}
\end{figure}

The results of Figure 5 in the main text showed that \textsc{Waldo} is able to leverage an estimated posterior to construct conditionally valid confidence regions, even when the prior is at odds with the data. On the other side, when no prior information is available, it is common to sample $\btheta$ according to a uniform distribution over the parameter space. In this case, we observe that confidence sets and posterior credible regions largely overlap. Nonetheless, if the latter happen to suffer from approximation errors, as is common for neural posteriors in high dimensions, this could hinder the statistical reliability of the estimated region. \textsc{Waldo} can correct even for this problem and guarantee conditional coverage, as we can see from panel \textit{a)} in Figure~\ref{fig:sets_and_area}.

\begin{figure}[!t]
    %\floatbox[{\capbeside\thisfloatsetup{capbesideposition={left,top}}}]{figure}[\FBwidth]
    {\caption{\small \textbf{Coverage diagnostics for Gaussian mixture model example with uniform prior}. We achieve correct conditional coverage for \textsc{Waldo} (left figure) but not for credible regions (right figure) even though the prior is is uniform, due to estimation and approximation errors, which  \textsc{Waldo} can correct via recalibration.}
    \label{fig:coverage_diagnostics_mixture_uniform}}
    {\includegraphics[width=0.7\textwidth]{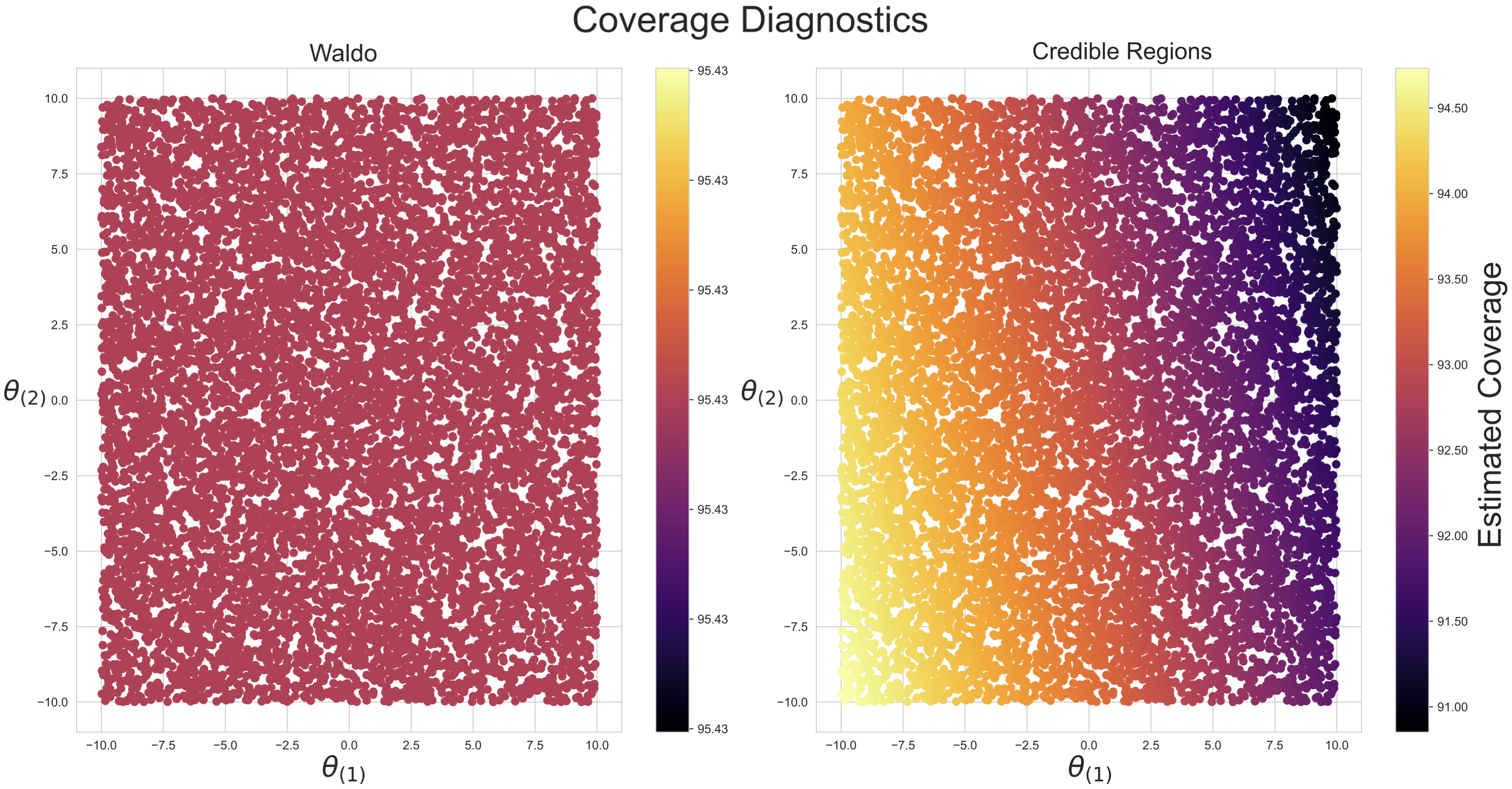}}
\end{figure}

Figure~\ref{fig:coverage_diagnostics_mixture_uniform} shows the output of the diagnostics procedure when using a uniform prior to train the posterior estimator (compare with Figure 5, right column, in the main text, which used a Gaussian prior). We achieve correct conditional coverage for \textsc{Waldo} but not for credible regions even though the prior is is uniform, due to estimation and approximation errors in the posterior, which \textsc{Waldo} can correct using quantile regression to calibrate the test statistics.

\subsection{Confidence Sets for Muon Energies using CNN
Predictions}
\label{sec: additional_muons}

\begin{figure}[!b]
    % \floatbox[{\capbeside\thisfloatsetup{capbesideposition={right,top},capbesidewidth=0.15\textwidth}}]{figure}[\FBwidth]
    {\caption{\small \textbf{Confidence and prediction sets for the muon energy reconstruction experiment}. Boxplots of the upper and lower bounds of prediction sets (green) versus \textsc{Waldo} confidence sets (red) for full the calorimeter data, all divided in 19 bins over true energy. We clearly see the bias occurring in the prediction sets (especially at high energies) and the correction applied by \textsc{Waldo}.}
    \label{fig:muons68_prediction_comparison}}
    {\includegraphics[width=0.5\textwidth]{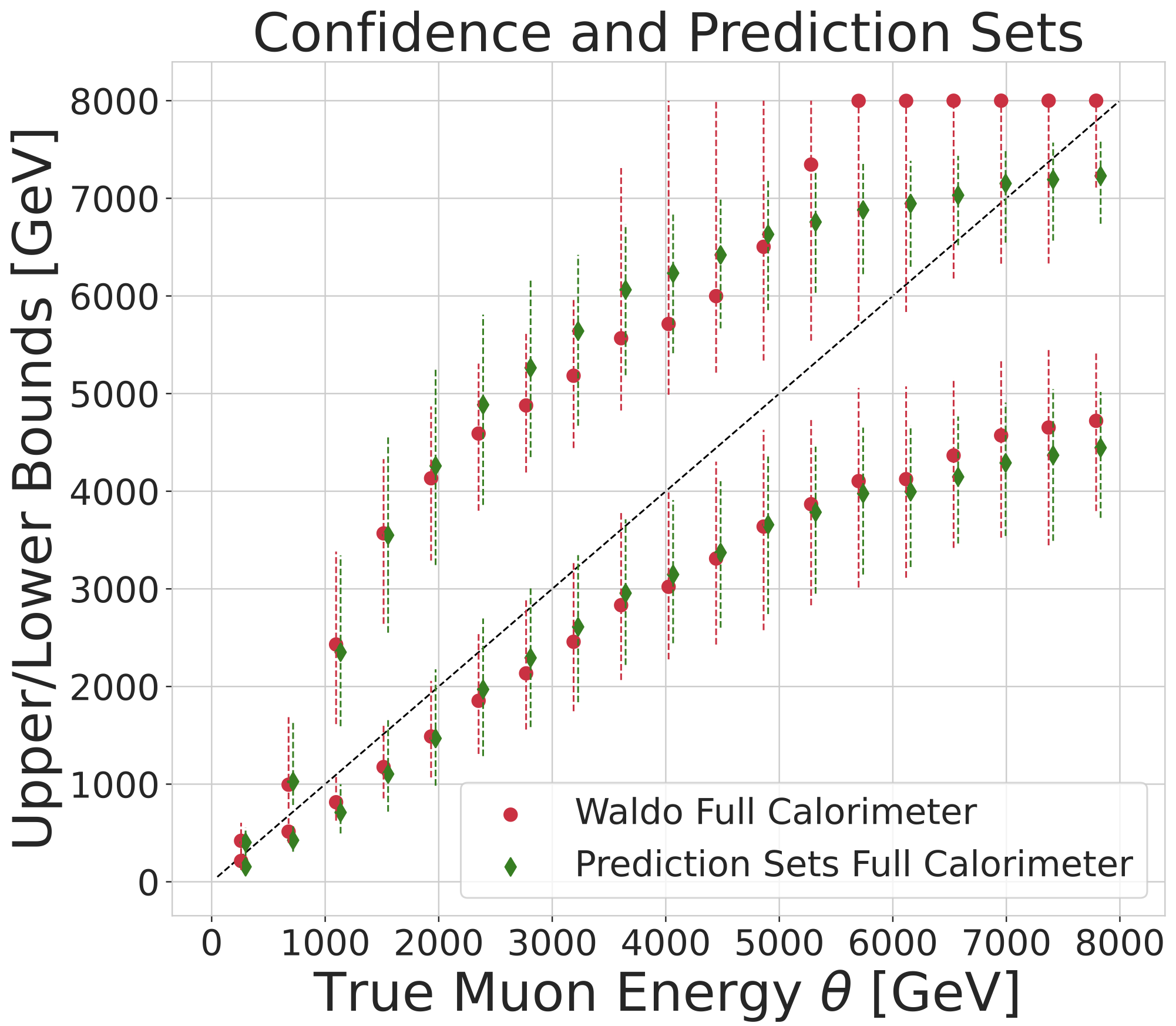}}
\end{figure}

Figure~\ref{fig:muons68_prediction_comparison} compares confidence sets and prediction sets for the full calorimeter data, showing clearly the bias in the prediction sets and the correction applied by Waldo. These results explain the observed patterns in Figure 6 in the main text: prediction sets are centered around the point prediction, which is downward biased at high energies, mainly due to the nonlinearity of the response at high energies.

\section{DETAILS ON MODELS, TRAINING, AND COMPUTATIONAL RESOURCES}

\subsection{Synthetic Examples for Statistical Properties}
\label{sec: details_stat_properties}

See Section~\ref{sec: waldo_stat_properties} in the main text and Appendix~\ref{sec: property3} for descriptions of the experiments. For \textsc{Property I} and \textsc{Property II}, we used the implementation of local linear regression available in \cite{Seabold2010StatsmodelsEA} to estimate conditional mean and conditional variance within a prediction setting, with $B=20{,}000$. For \textsc{Property III}, instead, we used a simple neural network with one hidden layer and $B=50{,}000$. In all cases, for quantile regression we used quantile Gradient Boosting \cite{pedregosa2011scikit}, with $B^\prime=20{,}000$ for \textsc{Property I} and \textsc{Property II}, and $B^\prime=50{,}000$ for \textsc{Property III}. All models were trained on a MacBook Pro M1Pro (CPU only). 

\subsection{Synthetic Example for Computational Properties}
\label{sec: details_comput_properties}

See Section~\ref{sec: comput_properties} in the main text for a description of the experiment. To compute the test statistic $\tau^{\textsc{Waldo}}$, we approximated conditional mean and conditional variance through a posterior distribution estimated via normalizing flows     \citep{tejero-cantero2020sbi}, with $B=20{,}000$ for $p=1$ and  $B=200{,}000$ for $p=10$. To construct the confidence sets, critical values were then estimated both via quantile regression using quantile Gradient Boosting \citep{pedregosa2011scikit} with varying values of $B^\prime$, and via Monte Carlo by simulating many times for each $\btheta$ and retaining the $(1-\alpha)$ quantile of the computed test statistics. The evaluation set was made of $1{,}000$ samples over $\bTheta = [-1, 1]^p$. To make the comparison fair, if quantile regression used $B^\prime=50{,}000$, then Monte Carlo had access to $50$ simulations for each of the $1{,}000$ samples in the evaluation set. The estimated coverage probability for both methods was then estimated using the implementation of Generalized Additive Models (GAMs) with thin plate splines available in the \texttt{MGCV} package of \texttt{R}, with $B^{\prime\prime}=30{,}000$.

\subsection{Confidence Sets from Neural Posteriors: Two-Dimensional Gaussian Mixture}
\label{sec: details_neural_posterior}

See Section~\ref{sec: mixture_bayesianLFI} in the main text and Appendix~\ref{sec: additional_neural_post} for descriptions of the experiments and details on the algorithms and sample sizes used. Training was done on a MacBook Pro M1Pro (CPU only); it took approximately 15--20 minutes to train the posterior estimator, and an additional $\sim$2 minutes for the quantile neural network to estimate the critical values. Note that the latter step requires computing the conditional mean, the conditional variance and the Waldo statistic over all sample points in $\mathcal{T}'$. 
The posterior was sampled multiple times for each $\x \in \mathcal{T}'$ to approximate $\mathbb{E}(\btheta|\x)$ and $\mathbb{V}(\btheta|\x)$ via Monte Carlo; this procedure took a total of $\sim$45 minutes (but could potentially be optimized through vectorizations in the future). 

\subsection{Confidence Sets for Muon Energies using CNN
Predictions}
\label{sec: details_muons}

See Section~\ref{sec: muons} and Appendix~\ref{sec: additional_muons} for descriptions of the experiment and details on the algorithms and sample sizes used. We had access to 886,716 simulated muons in total; roughly 200,000 muons were used to estimate the critical values, $\sim$24,000 muons to construct the final confidence sets and diagnostics, and the rest was used to estimate the conditional mean and variance via the custom 3D CNN from \citet{kieseler2022calorimetric}. Training the latter CNN took approximately 20 hours for the conditional mean and another 20 hours for the conditional variance, using an NVIDIA V100 GPU on an Azure cloud computing machine. Estimating the critical values via quantile gradient boosted trees took approximately 2 minutes.

%% file: main.bbl
\begin{thebibliography}{55}
\providecommand{\natexlab}[1]{#1}
\providecommand{\url}[1]{\texttt{#1}}
\expandafter\ifx\csname urlstyle\endcsname\relax
  \providecommand{\doi}[1]{doi: #1}\else
  \providecommand{\doi}{doi: \begingroup \urlstyle{rm}\Url}\fi

\bibitem[Aad et~al.(2012)Aad, Abajyan, Abbott, Abdallah, Khalek, Abdelalim,
  Aben, Abi, Abolins, AbouZeid, et~al.]{aad2012observation}
Georges Aad, Tatevik Abajyan, B~Abbott, J~Abdallah, S~Abdel Khalek, Ahmed~Ali
  Abdelalim, R~Aben, B~Abi, M~Abolins, OS~AbouZeid, et~al.
\newblock Observation of a new particle in the search for the {S}tandard
  {M}odel {H}iggs boson with the {ATLAS} detector at the {LHC}.
\newblock \emph{Physics Letters B}, 716\penalty0 (1):\penalty0 1--29, 2012.

\bibitem[Agostinelli et~al.(2003)Agostinelli, Allison, Amako, Apostolakis,
  Araujo, Arce, Asai, Axen, Banerjee, Barrand, et~al.]{agostinelli2003geant4}
Sea Agostinelli, John Allison, K~Amako, John Apostolakis, H~Araujo, Pedro Arce,
  Makoto Asai, D~Axen, Swagato Banerjee, G~Barrand, et~al.
\newblock {GEANT4}—a simulation toolkit.
\newblock \emph{Nuclear Instruments and Methods in Physics Research A},
  506\penalty0 (3):\penalty0 250--303, 2003.

\bibitem[Augustin et~al.(1974)Augustin, Boyarski, Breidenbach, Bulos, Dakin,
  Feldman, Fischer, Fryberger, Hanson, Jean-Marie,
  et~al.]{augustin1974discovery}
J-E Augustin, Adam~M Boyarski, Martin Breidenbach, F~Bulos, JT~Dakin,
  GJ~Feldman, GE~Fischer, D~Fryberger, G~Hanson, B~Jean-Marie, et~al.
\newblock Discovery of a narrow resonance in e+ e- annihilation.
\newblock \emph{Physical Review Letters}, 33\penalty0 (23):\penalty0 1406,
  1974.

\bibitem[Bayarri and Berger(2004)]{Bayarri2004}
M.~J. Bayarri and J.~O. Berger.
\newblock The interplay of {B}ayesian and frequentist analysis.
\newblock \emph{Statistical Science}, 19\penalty0 (1):\penalty0 58--80, 2004.
\newblock \doi{10.1214/088342304000000116}.

\bibitem[Berger(2006)]{Berger2006}
James Berger.
\newblock {The case for objective {B}ayesian analysis}.
\newblock \emph{Bayesian Analysis}, 1\penalty0 (3):\penalty0 385--402, 2006.
\newblock \doi{10.1214/06-BA115}.

\bibitem[Bordoloi et~al.(2010)Bordoloi, Lilly, and Amara]{bordoloi2010photo}
Rongmon Bordoloi, Simon~J Lilly, and Adam Amara.
\newblock Photo-z performance for precision cosmology.
\newblock \emph{Monthly Notices of the Royal Astronomical Society},
  406\penalty0 (2):\penalty0 881--895, 2010.

\bibitem[Boyda et~al.(2021)Boyda, Kanwar, Racani{\`e}re, Rezende, Albergo,
  Cranmer, Hackett, and Shanahan]{boyda2021sampling}
Denis Boyda, Gurtej Kanwar, S{\'e}bastien Racani{\`e}re, Danilo~Jimenez
  Rezende, Michael~S Albergo, Kyle Cranmer, Daniel~C Hackett, and Phiala~E
  Shanahan.
\newblock Sampling using su (n) gauge equivariant flows.
\newblock \emph{Physical Review D}, 103\penalty0 (7):\penalty0 074504, 2021.

\bibitem[Brehmer et~al.(2020)Brehmer, Louppe, Pavez, and Cranmer]{Brehmer2020}
Johann Brehmer, Gilles Louppe, Juan Pavez, and Kyle Cranmer.
\newblock Mining gold from implicit models to improve likelihood-free
  inference.
\newblock \emph{Proceedings of the National Academy of Sciences}, 117\penalty0
  (10):\penalty0 5242--5249, 2020.
\newblock \doi{10.1073/pnas.1915980117}.

\bibitem[{{CDF} Collaboration}(1995)]{cdf1995observation}
The {{CDF} Collaboration}.
\newblock Observation of top quark production in {Pbar-P} collisions.
\newblock \emph{arXiv preprint hep-ex/9503002}, 1995.

\bibitem[Chao(1970)]{chao1970asymptotic}
MT~Chao.
\newblock The asymptotic behavior of {B}ayes' estimators.
\newblock \emph{The Annals of Mathematical Statistics}, 41\penalty0
  (2):\penalty0 601--608, 1970.

\bibitem[Chatrchyan et~al.(2012)Chatrchyan, Khachatryan, Sirunyan, Tumasyan,
  Adam, Aguilo, Bergauer, Dragicevic, Er{\"o}, Fabjan,
  et~al.]{chatrchyan2012observation}
Serguei Chatrchyan, Vardan Khachatryan, Albert~M Sirunyan, Armen Tumasyan,
  Wolfgang Adam, Ernest Aguilo, Thomas Bergauer, M~Dragicevic, J~Er{\"o},
  C~Fabjan, et~al.
\newblock Observation of a new boson at a mass of 125 gev with the cms
  experiment at the lhc.
\newblock \emph{Physics Letters B}, 716\penalty0 (1):\penalty0 30--61, 2012.

\bibitem[Chen and Guestrin(2016)]{chen2016xgboost}
Tianqi Chen and Carlos Guestrin.
\newblock Xgboost: A scalable tree boosting system.
\newblock In \emph{Proceedings of the 22nd acm sigkdd international conference
  on knowledge discovery and data mining}, pages 785--794, 2016.

\bibitem[Chernozhukov et~al.(2021)Chernozhukov, W{\"u}thrich, and
  Zhu]{chernozhukov2021distributional}
Victor Chernozhukov, Kaspar W{\"u}thrich, and Yinchu Zhu.
\newblock Distributional conformal prediction.
\newblock \emph{Proceedings of the National Academy of Sciences}, 118\penalty0
  (48):\penalty0 e2107794118, 2021.

\bibitem[Clart{\'e} et~al.(2021)Clart{\'e}, Robert, Ryder, and
  Stoehr]{clarte2021componentwise}
Gr{\'e}goire Clart{\'e}, Christian~P Robert, Robin~J Ryder, and Julien Stoehr.
\newblock Componentwise approximate {B}ayesian computation via {G}ibbs-like
  steps.
\newblock \emph{Biometrika}, 108\penalty0 (3):\penalty0 591--607, 2021.

\bibitem[Cousins(2018)]{cousins2018lectures}
Robert~D. Cousins.
\newblock Lectures on statistics in theory: Prelude to statistics in practice,
  2018.
\newblock URL \url{https://arxiv.org/abs/1807.05996}.

\bibitem[Cranmer et~al.(2020)Cranmer, Brehmer, and Louppe]{cranmer2020frontier}
Kyle Cranmer, Johann Brehmer, and Gilles Louppe.
\newblock The frontier of simulation-based inference.
\newblock \emph{Proceedings of the National Academy of Sciences}, 117\penalty0
  (48):\penalty0 30055--30062, 2020.

\bibitem[Dalmasso et~al.(2021)Dalmasso, Masserano, Zhao, Izbicki, and
  Lee]{dalmasso2021likelihood}
Niccolo Dalmasso, Luca Masserano, David Zhao, Rafael Izbicki, and Ann~B Lee.
\newblock {Likelihood-Free Frequentist Inference: Confidence Sets with Correct
  Conditional Coverage}.
\newblock \emph{arXiv preprint arXiv:2107.03920}, 2021.

\bibitem[Datta and Sweeting(2005)]{Datta2005}
Gauri~Sankar Datta and Trevor~J. Sweeting.
\newblock Probability matching priors.
\newblock In D.K. Dey and C.R. Rao, editors, \emph{Bayesian Thinking},
  volume~25 of \emph{Handbook of Statistics}, pages 91--114. Elsevier, 2005.
\newblock \doi{https://doi.org/10.1016/S0169-7161(05)25003-4}.

\bibitem[Dey et~al.(2022)Dey, Zhao, Newman, Andrews, Izbicki, and
  Lee]{dey2022calibrated}
Biprateep Dey, David Zhao, Jeffrey~A Newman, Brett~H Andrews, Rafael Izbicki,
  and Ann~B Lee.
\newblock Calibrated predictive distributions via diagnostics for conditional
  coverage.
\newblock \emph{arXiv preprint arXiv:2205.14568}, 2022.

\bibitem[Dorigo et~al.(2022)Dorigo, Guglielmini, Kieseler, Layer, and
  Strong]{dorigo2022deep}
Tommaso Dorigo, Sofia Guglielmini, Jan Kieseler, Lukas Layer, and Giles~C
  Strong.
\newblock Deep regression of muon energy with a k-nearest neighbor algorithm.
\newblock \emph{arXiv preprint arXiv:2203.02841}, 2022.

\bibitem[Durkan et~al.(2020)Durkan, Bekasov, Murray, and Papamakarios]{nflows}
Conor Durkan, Artur Bekasov, Iain Murray, and George Papamakarios.
\newblock {\texttt{nflows}}: {N}ormalizing flows in {PyTorch}, November 2020.
\newblock URL \url{https://doi.org/10.5281/zenodo.4296287}.

\bibitem[Gal and Ghahramani(2016)]{Gal2016}
Yarin Gal and Zoubin Ghahramani.
\newblock Dropout as a {B}ayesian approximation: Representing model uncertainty
  in deep learning.
\newblock In Maria~Florina Balcan and Kilian~Q. Weinberger, editors,
  \emph{Proceedings of The 33rd International Conference on Machine Learning},
  volume~48 of \emph{Proceedings of Machine Learning Research}, pages
  1050--1059, New York, New York, USA, 20--22 Jun 2016. PMLR.
\newblock URL \url{https://proceedings.mlr.press/v48/gal16.html}.

\bibitem[Gawlikowski et~al.(2021)Gawlikowski, Tassi, Ali, Lee, Humt, Feng,
  Kruspe, Triebel, Jung, Roscher, et~al.]{gawlikowski2021survey}
Jakob Gawlikowski, Cedrique Rovile~Njieutcheu Tassi, Mohsin Ali, Jongseok Lee,
  Matthias Humt, Jianxiang Feng, Anna Kruspe, Rudolph Triebel, Peter Jung,
  Ribana Roscher, et~al.
\newblock A survey of uncertainty in deep neural networks.
\newblock \emph{arXiv preprint arXiv:2107.03342}, 2021.

\bibitem[Gerber and Nychka(2021)]{gerber2021fast}
Florian Gerber and Douglas Nychka.
\newblock Fast covariance parameter estimation of spatial {G}aussian process
  models using neural networks.
\newblock \emph{Stat}, 10\penalty0 (1):\penalty0 e382, 2021.

\bibitem[Ghosh(1991)]{ghosh1991higher}
JK~Ghosh.
\newblock Higher order asymptotics for the likelihood ratio, rao's and wald's
  tests.
\newblock \emph{Statistics \& probability letters}, 12\penalty0 (6):\penalty0
  505--509, 1991.

\bibitem[Ghosh and Ramamoorthi(2003)]{ghosh2003preliminaries}
JK~Ghosh and RV~Ramamoorthi.
\newblock Preliminaries and the finite dimensional case.
\newblock \emph{Bayesian Nonparametrics}, pages 9--55, 2003.

\bibitem[Ghosh et~al.(1982)Ghosh, Sinha, and Joshi]{ghosh1982expansions}
JK~Ghosh, BK~Sinha, and SN~Joshi.
\newblock Expansions for posterior probability and integrated {B}ayes risk.
\newblock \emph{Statistical Decision Theory and Related Topics III},
  1:\penalty0 403--456, 1982.

\bibitem[Herb et~al.(1977)Herb, Hom, Lederman, Sens, Snyder, Yoh, Appel, Brown,
  Brown, Innes, et~al.]{herb1977observation}
SW~Herb, DC~Hom, LM~Lederman, JC~Sens, HD~Snyder, JK~Yoh, JA~Appel, BC~Brown,
  CN~Brown, WR~Innes, et~al.
\newblock Observation of a dimuon resonance at 9.5 {GeV} in 400-{GeV}
  proton-nucleus collisions.
\newblock \emph{Physical Review Letters}, 39\penalty0 (5):\penalty0 252, 1977.

\bibitem[Hermans et~al.(2021)Hermans, Delaunoy, Rozet, Wehenkel, and
  Louppe]{hermans2021averting}
Joeri Hermans, Arnaud Delaunoy, Fran{\c{c}}ois Rozet, Antoine Wehenkel, and
  Gilles Louppe.
\newblock Averting a crisis in simulation-based inference.
\newblock \emph{arXiv preprint arXiv:2110.06581}, 2021.

\bibitem[Ho et~al.(2019)Ho, Rau, Ntampaka, Farahi, Trac, and
  P{\'o}czos]{ho2019robust}
Matthew Ho, Markus~Michael Rau, Michelle Ntampaka, Arya Farahi, Hy~Trac, and
  Barnab{\'a}s P{\'o}czos.
\newblock A robust and efficient deep learning method for dynamical mass
  measurements of galaxy clusters.
\newblock \emph{The Astrophysical Journal}, 887\penalty0 (1):\penalty0 25,
  2019.

\bibitem[Kass and Wasserman(1996)]{Kass1996}
Robert~E. Kass and Larry Wasserman.
\newblock The selection of prior distributions by formal rules.
\newblock \emph{Journal of the American Statistical Association}, 91\penalty0
  (435):\penalty0 1343--1370, 1996.
\newblock \doi{10.1080/01621459.1996.10477003}.

\bibitem[Kiel et~al.(2019)Kiel, O'Dell, Fisher, Eldering, Nassar, MacDonald,
  and Wennberg]{kiel2019bias}
Matth{\"a}us Kiel, Christopher~W O'Dell, Brendan Fisher, Annmarie Eldering, Ray
  Nassar, Cameron~G MacDonald, and Paul~O Wennberg.
\newblock How bias correction goes wrong: Measurement of {$X_{CO_2}$} affected
  by erroneous surface pressure estimates.
\newblock \emph{Atmospheric Measurement Techniques}, 12\penalty0 (4):\penalty0
  2241--2259, 2019.

\bibitem[Kieseler et~al.(2021)Kieseler, Strong, Chiandotto, Dorigo, and
  Layer]{kieseler_jan_2021_5163817}
Jan Kieseler, Giles~Chatham Strong, Filippo Chiandotto, Tommaso Dorigo, and
  Lukas Layer.
\newblock Preprocessed dataset for ``{C}alorimetric measurement of multi-{TeV}
  muons via deep regression", August 2021.
\newblock URL \url{https://doi.org/10.5281/zenodo.5163817}.

\bibitem[Kieseler et~al.(2022)Kieseler, Strong, Chiandotto, Dorigo, and
  Layer]{kieseler2022calorimetric}
Jan Kieseler, Giles~C Strong, Filippo Chiandotto, Tommaso Dorigo, and Lukas
  Layer.
\newblock Calorimetric measurement of multi-{TeV} muons via deep regression.
\newblock \emph{The European Physical Journal C}, 82\penalty0 (1):\penalty0
  1--26, 2022.

\bibitem[LeCun et~al.(1995)LeCun, Bengio, et~al.]{lecun1995convolutional}
Yann LeCun, Yoshua Bengio, et~al.
\newblock Convolutional networks for images, speech, and time series.
\newblock \emph{The handbook of brain theory and neural networks},
  3361\penalty0 (10):\penalty0 1995, 1995.

\bibitem[Lehmann et~al.(2005)Lehmann, Romano, and Casella]{lehmann2005testing}
Erich~Leo Lehmann, Joseph~P Romano, and George Casella.
\newblock \emph{Testing statistical hypotheses}, volume~3.
\newblock Springer, 2005.

\bibitem[Lei et~al.(2018)Lei, G’Sell, Rinaldo, Tibshirani, and
  Wasserman]{lei2018distribution}
Jing Lei, Max G’Sell, Alessandro Rinaldo, Ryan~J Tibshirani, and Larry
  Wasserman.
\newblock Distribution-free predictive inference for regression.
\newblock \emph{Journal of the American Statistical Association}, 113\penalty0
  (523):\penalty0 1094--1111, 2018.

\bibitem[Li et~al.(2020)Li, Yu, and Zeng]{li2020deviance}
Yong Li, Jun Yu, and Tao Zeng.
\newblock Deviance information criterion for latent variable models and
  misspecified models.
\newblock \emph{Journal of Econometrics}, 216\penalty0 (2):\penalty0 450--493,
  2020.

\bibitem[Lueckmann et~al.(2021)Lueckmann, Boelts, Greenberg, Goncalves, and
  Macke]{lueckmann2021benchmarking}
Jan-Matthis Lueckmann, Jan Boelts, David Greenberg, Pedro Goncalves, and Jakob
  Macke.
\newblock Benchmarking simulation-based inference.
\newblock In \emph{International Conference on Artificial Intelligence and
  Statistics}, pages 343--351. PMLR, 2021.

\bibitem[Meinshausen and Ridgeway(2006)]{meinshausen2006quantile}
Nicolai Meinshausen and Greg Ridgeway.
\newblock Quantile regression forests.
\newblock \emph{Journal of Machine Learning Research}, 7\penalty0 (6), 2006.

\bibitem[Mishra-Sharma and Cranmer(2022)]{mishra2022neural}
Siddharth Mishra-Sharma and Kyle Cranmer.
\newblock Neural simulation-based inference approach for characterizing the
  galactic center $\gamma$-ray excess.
\newblock \emph{Physical Review D}, 105\penalty0 (6):\penalty0 063017, 2022.

\bibitem[Neyman(1937)]{neyman1937outline}
Jerzy Neyman.
\newblock Outline of a theory of statistical estimation based on the classical
  theory of probability.
\newblock \emph{Philosophical Transactions of the Royal Society of London.
  Series A, Mathematical and Physical Sciences}, 236\penalty0 (767):\penalty0
  333--380, 1937.

\bibitem[Papadopoulos et~al.(2007)Papadopoulos, Vovk, and
  Gammerman]{papadopoulos2007conformal}
Harris Papadopoulos, Volodya Vovk, and Alex Gammerman.
\newblock Conformal prediction with neural networks.
\newblock In \emph{19th IEEE International Conference on Tools with Artificial
  Intelligence (ICTAI 2007)}, volume~2, pages 388--395. IEEE, 2007.

\bibitem[Papamakarios et~al.(2021)Papamakarios, Nalisnick, Rezende, Mohamed,
  and Lakshminarayanan]{papamakarios2021normalizing}
George Papamakarios, Eric Nalisnick, Danilo~Jimenez Rezende, Shakir Mohamed,
  and Balaji Lakshminarayanan.
\newblock Normalizing flows for probabilistic modeling and inference.
\newblock \emph{Journal of Machine Learning Research}, 22\penalty0
  (57):\penalty0 1--64, 2021.

\bibitem[Patil et~al.(2022)Patil, Kuusela, and Hobbs]{patil2020objective}
Pratik Patil, Mikael Kuusela, and Jonathan Hobbs.
\newblock Objective frequentist uncertainty quantification for atmospheric
  $\mathrm{CO}_2$ retrievals.
\newblock \emph{SIAM/ASA Journal on Uncertainty Quantification}, 10\penalty0
  (3):\penalty0 827--859, 2022.
\newblock \doi{10.1137/20M1356403}.

\bibitem[Pedregosa et~al.(2011)Pedregosa, Varoquaux, Gramfort, Michel, Thirion,
  Grisel, Blondel, Prettenhofer, Weiss, Dubourg, et~al.]{pedregosa2011scikit}
Fabian Pedregosa, Ga{\"e}l Varoquaux, Alexandre Gramfort, Vincent Michel,
  Bertrand Thirion, Olivier Grisel, Mathieu Blondel, Peter Prettenhofer, Ron
  Weiss, Vincent Dubourg, et~al.
\newblock Scikit-learn: Machine learning in python.
\newblock \emph{the Journal of machine Learning research}, 12:\penalty0
  2825--2830, 2011.

\bibitem[Scricciolo(1999)]{Scricciolo1999}
Catia Scricciolo.
\newblock Probability matching priors: A review.
\newblock \emph{Journal of the Italian Statistical Society}, 8:\penalty0
  83--100, 1999.
\newblock \doi{10.1007/BF03178943}.

\bibitem[Seabold and Perktold(2010)]{Seabold2010StatsmodelsEA}
Skipper Seabold and Josef Perktold.
\newblock Statsmodels: Econometric and statistical modeling with python.
\newblock 2010.

\bibitem[Simola et~al.(2021)Simola, Cisewski-Kehe, Gutmann, and
  Corander]{simola2021adaptive}
Umberto Simola, Jessi Cisewski-Kehe, Michael~U Gutmann, and Jukka Corander.
\newblock Adaptive {A}pproximate {B}ayesian {C}omputation tolerance selection.
\newblock \emph{Bayesian analysis}, 16\penalty0 (2):\penalty0 397--423, 2021.

\bibitem[Sisson et~al.(2007)Sisson, Fan, and Tanaka]{sisson2007sequential}
Scott~A Sisson, Yanan Fan, and Mark~M Tanaka.
\newblock Sequential {M}onte {C}arlo without likelihoods.
\newblock \emph{Proceedings of the National Academy of Sciences}, 104\penalty0
  (6):\penalty0 1760--1765, 2007.

\bibitem[Tejero-Cantero et~al.(2020)Tejero-Cantero, Boelts, Deistler,
  Lueckmann, Durkan, Gonçalves, Greenberg, and Macke]{tejero-cantero2020sbi}
Alvaro Tejero-Cantero, Jan Boelts, Michael Deistler, Jan-Matthis Lueckmann,
  Conor Durkan, Pedro~J. Gonçalves, David~S. Greenberg, and Jakob~H. Macke.
\newblock {\texttt{sbi}}: {A} toolkit for simulation-based inference.
\newblock \emph{Journal of Open Source Software}, 5\penalty0 (52):\penalty0
  2505, 2020.
\newblock \doi{10.21105/joss.02505}.
\newblock URL \url{https://doi.org/10.21105/joss.02505}.

\bibitem[Toni et~al.(2009)Toni, Welch, Strelkowa, Ipsen, and
  Stumpf]{toni2009approximate}
Tina Toni, David Welch, Natalja Strelkowa, Andreas Ipsen, and Michael~PH
  Stumpf.
\newblock Approximate {B}ayesian {C}omputation scheme for parameter inference
  and model selection in dynamical systems.
\newblock \emph{Journal of the Royal Society Interface}, 6\penalty0
  (31):\penalty0 187--202, 2009.

\bibitem[Vovk et~al.(2005)Vovk, Gammerman, and Shafer]{vovk2005algorithmic}
Vladimir Vovk, Alexander Gammerman, and Glenn Shafer.
\newblock \emph{Algorithmic learning in a random world}.
\newblock Springer Science \& Business Media, 2005.

\bibitem[Wald(1943)]{wald1943tests}
Abraham Wald.
\newblock Tests of statistical hypotheses concerning several parameters when
  the number of observations is large.
\newblock \emph{Transactions of the American Mathematical society}, 54\penalty0
  (3):\penalty0 426--482, 1943.

\bibitem[Wilks(1938)]{wilks1938LRAsymptotic}
S.~S. Wilks.
\newblock The large-sample distribution of the likelihood ratio for testing
  composite hypotheses.
\newblock \emph{Ann. Math. Statist.}, 9\penalty0 (1):\penalty0 60--62, 03 1938.
\newblock \doi{10.1214/aoms/1177732360}.

\end{thebibliography}
